# Patient-Specific Dynamic Digital-Physical Twin for Coronary Intervention Training: An Integrated Mixed Reality Approach

Shuo Wang [1,†], Tong Ren [2,†], Nan Cheng [2], Rong Wang [2,*], and Li Zhang [1,*]

[1] Department of Engineering Physics, Key Laboratory of Particle and Radiation Imaging, Ministry of Education, Tsinghua University, Beijing, China; zli@mail.tsinghua.edu.cn(L.Z.); shuo-wan19@mails.tsinghua.edu.cn(S.W.)

[2] Department of Adult Cardiac Surgery, Senior Department of Cardiology, The Six Medical Center of PLA General Hospital. Fucheng Road, Haidian District,100048, Beijing, China; wangrongd@126.com(R.W.); cn86919@163.com(N.C.); rentong8713@163.com(T.R.)

* Correspondence: wangrongd@126.com(R.W.); zli@mail.tsinghua.edu.cn(L.Z.)
† These authors contributed equally to this work.

## Abstract

**Background and Objective:** Precise preoperative planning and effective physician training for coronary interventions are increasingly important. Despite advances in medical imaging technologies, transforming static or limited dynamic imaging data into comprehensive dynamic cardiac models remains challenging. Existing training systems lack accurate simulation of cardiac physiological dynamics. This study develops a comprehensive dynamic cardiac model research framework based on 4D-CTA, integrating digital twin technology, computer vision, and physical model manufacturing to provide precise, personalized tools for interventional cardiology.

**Methods:** Using 4D-CTA data from a 60-year-old female with three-vessel coronary stenosis, we segmented cardiac chambers and coronary arteries, constructed dynamic models, and implemented skeletal skinning weight computation to simulate vessel deformation across 20 cardiac phases. Transparent vascular physical models were manufactured using medical-grade silicone. We developed cardiac output analysis and virtual angiography systems, implemented guidewire 3D reconstruction using binocular stereo vision, and evaluated the system through angiography validation and CABG training applications.

**Results:** Morphological consistency between virtual and real angiography reached 80.9%. Dice similarity coefficients for guidewire motion ranged from 0.741-0.812, with mean trajectory errors below 1.1 mm. The transparent model demonstrated advantages in CABG training, allowing direct visualization while simulating beating heart challenges.

**Conclusion:** Our patient-specific digital-physical twin approach effectively reproduces both anatomical structures and dynamic characteristics of coronary vasculature, offering a dynamic environment with visual and tactile feedback valuable for education and clinical planning.

## 1. Introduction

With the rapid development of interventional cardiology, precise preoperative planning and effective physician training have become increasingly important. Although modern medical imaging technologies such as CT angiography (CTA), magnetic resonance imaging (MRI), and X-ray angiography (XA) have provided powerful support for the diagnosis and treatment of cardiovascular diseases [1-3], transforming these static or limited dynamic imaging data into comprehensive dynamic cardiac models still faces many challenges. In particular, existing training systems largely lack accurate simulation of cardiac physiological dynamic characteristics and cannot truly reproduce the complex motion patterns of coronary arteries throughout the cardiac cycle [4,5].

In recent years, digital twin technology has attracted widespread attention in the medical field. This technology creates virtual replicas of physical entities, enabling real-time data exchange and state synchronization [6]. In cardiac intervention, digital twin technology has the potential to build personalized dynamic cardiac models to support precision medicine [7]. However, most current cardiac digital twin systems remain at the stage of static anatomical reconstruction or simplified dynamics simulation, lacking comprehensive capture of patient-specific dynamic characteristics [8,9].

Augmented reality (AR) and mixed reality (MR) technologies provide new visualization and interaction methods for interventional cardiology [10,11]. Annabestani et al. [12] explored the applications of AR/MR in interventional cardiology, noting that these technologies can enhance surgical navigation and medical education. However, existing AR/MR systems mostly rely on pre-acquired static data or simplified dynamic models, unable to accurately reflect patient-specific cardiac dynamic characteristics [13].

Furthermore, current interventional training systems mainly fall into two categories: computer-based virtual simulators and physical vascular models [14,15]. While virtual simulators provide good visual feedback, they lack realistic tactile experience; physical models provide tactile feedback but are typically static and unable to simulate cardiac pulsation and vascular deformation [16]. In particular, training systems for complex cardiac surgeries such as CABG lack dynamic models that can simultaneously provide visual and tactile feedback [17,18].

To address these issues, this study proposes a comprehensive dynamic cardiac model research framework based on 4D-CTA. This framework integrates digital twin technology, computer vision, and physical model manufacturing to achieve complete transformation from patient-specific imaging data to a dynamic physical-virtual combined training system. Compared to existing methods, our system has the following innovations: (1) high-precision dynamic coronary artery models constructed from real patient 4D-CTA data; (2) a comprehensive digital twin system integrating cardiac output analysis and virtual angiography; (3) a mixed reality training environment combining physical models and virtual imaging; (4) a technical pathway supporting guidewire 3D reconstruction and dynamic characteristic analysis.

This research aims to provide more precise, personalized preoperative planning and training tools for interventional cardiology through this comprehensive framework, thereby improving the safety and effectiveness of interventional treatments.

## 2. Materials and methods

This study establishes a comprehensive dynamic cardiac model research framework, as shown in Figure 1. The research begins with patient data acquisition (2.1), utilizing dynamic cardiac data with 11 temporal phases and a three-vessel coronary stenosis case as the foundation for subsequent work. In the digital twin development (2.2) phase, heart chamber segmentation, coronary artery segmentation, and basic structure segmentation are performed to construct dynamic coronary models and aortic valve reconstruction models. The research further branches into three key directions: physical twin fabrication (2.2.3), cardiac output analysis (2.3), and dynamic virtual angiography (2.4). The results from both physical models and virtual simulations provide essential support for guidewire 3D reconstruction (2.5). Finally, the entire research system is evaluated in the validation and applications section, including coronary angiography validation (3.1), guidewire motion characteristics analysis (3.2), and CABG surgical training (3.3). The chapter structure illustrated in Figure 1 clearly presents the complete research pathway from basic data to clinical applications, with each component closely interconnected, forming a systematic research methodology.

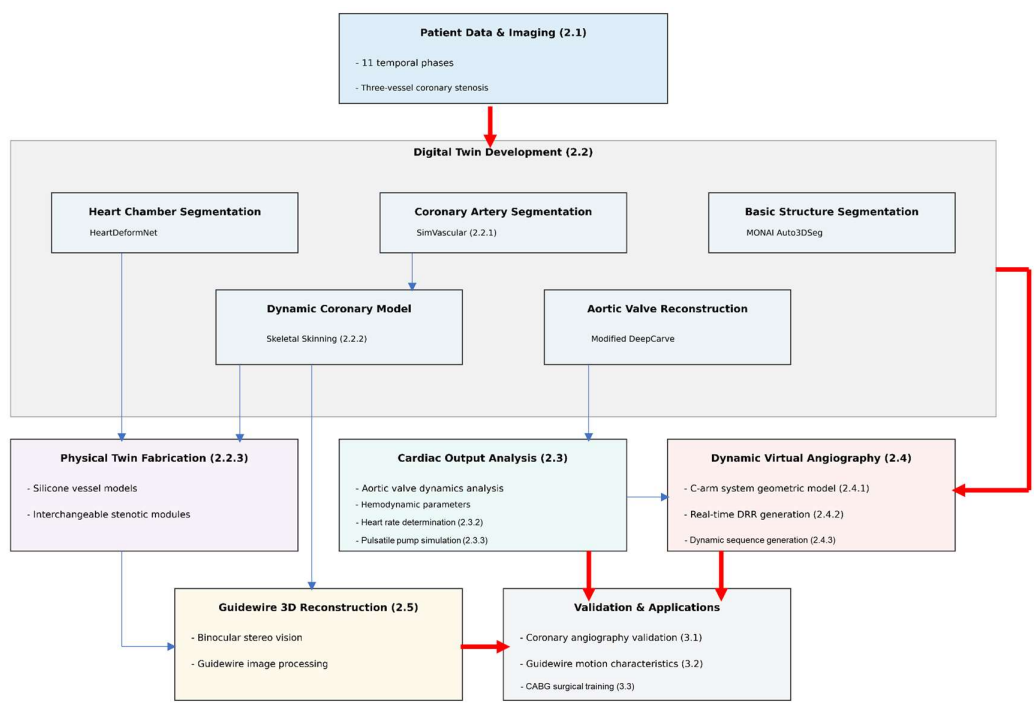

Figure 1: Workflow diagram of the dynamic cardiac model research framework, illustrating the progression from patient data acquisition through digital twin development to clinical validation and applications.

### 2.1. Patient Information and Imaging Protocol

A 60-year-old female patient with significant three-vessel coronary artery stenosis and no other known comorbidities was admitted to The Six Medical Center of PLA General Hospital. Following comprehensive evaluation, CABG with cardiopulmonary bypass (CPB) was deemed necessary. Preoperative imaging was performed using a GE APEX CT scanner (Revolution Apex, GE Healthcare, Milwaukee, MI) with images acquired during end-expiratory breath-hold and in sinus rhythm. The CT dataset was reconstructed into 11 temporal phases corresponding to -5% to 106% of the R-R interval. All images were reconstructed with 0.625 mm slice thickness, 0.625 mm increments, and 0.23 × 0.23 mm in-plane resolution. The mean effective radiation dose was 10.49 mSv, calculated according to European guidelines for multilayer CT.

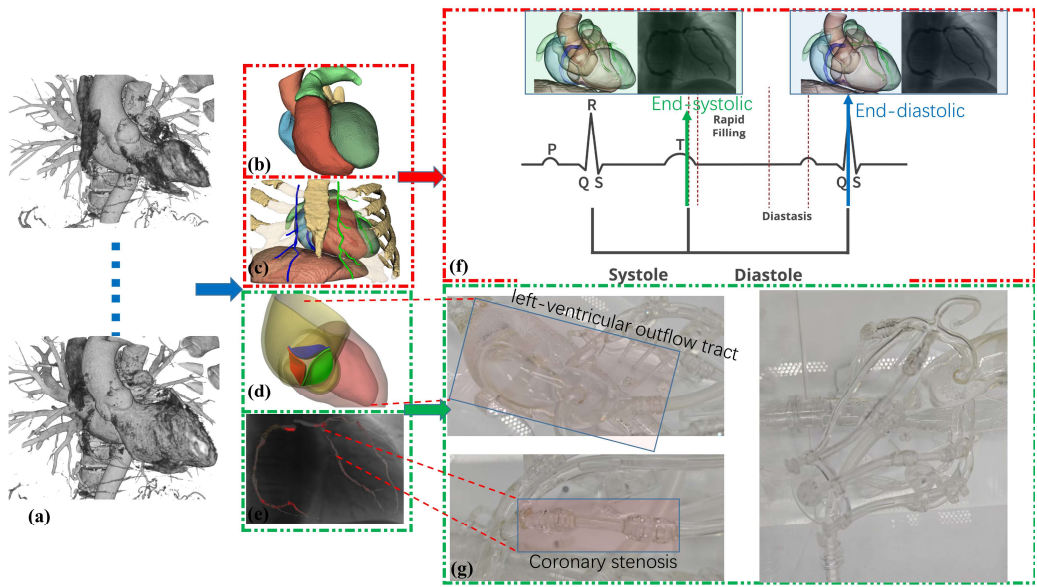

Figure 2: Comprehensive cardiac imaging and modeling workflow for digital and physical cardiovascular model development. The process begins with (a) cardiac-gated imaging sequences serving as foundational data, followed by sequential construction of (b) 3D heart chamber mesh models, (c) thoracic cage, diaphragm, and internal mammary artery models, (d) aortic valve models with left ventricular outflow tract visualization, (e) coronary artery models showing vessel structure, (f) dynamic cardiac cycle model sequences corresponding to systole and diastole phases with ECG correlation, and ultimately (g) physical twin model fabrication.

As shown in the Figure 2, the modeling of dynamic cardiovascular systems primarily comprises the following steps: (1) Cardiac chambers and great vessels segmentation: HeartDeformNet [19] was applied to automatically segment the left ventricle, right ventricle, left atrium, right atrium, left ventricular myocardium, aorta, and pulmonary artery across all phases of the cardiac cycle. (2) Basic structure segmentation and reconstruction: Based on the Auto3DSeg open-source project within the MONAI framework [20], we segmented and reconstructed the ribs, spine, sternum, diaphragm, and internal mammary artery from end-diastolic (70% phase) CT images. (3) Coronary artery segmentation: Due to severe stenosis in all three coronary arteries, manual annotation was used for vessel segmentation. (4) Dynamic coronary model construction: We implemented a skeletal skinning-based approach using biharmonic energy minimization to simulate coronary vessel deformation throughout 20 phases of the cardiac cycle. (5) Aortic valve reconstruction: A modified version of DeepCarve [21] was employed for aortic valve mesh reconstruction, with the key improvement being the utilization of the end-diastolic left ventricular model as a template, which features fewer distorted elements. For detailed processing methods of the complete workflow, readers can refer to our previous work [22].

*2.2. Physical-Digital Twin of Dynamic Coronary Arteries*

This section describes our comprehensive methodology for developing patient-specific coronary artery models that accurately replicate both anatomical and dynamic characteristics of the coronary vasculature. Our approach integrates advanced image processing techniques with novel biomechanical modeling to create a complete digital-physical twin system. Beginning with precise manual segmentation of severely stenotic coronary arteries, we then implement deformation algorithms to simulate cardiac-cycle-induced vessel motion. Finally, we translate these digital models into high-fidelity physical phantoms using specialized fabrication techniques. The resulting system provides an anatomically accurate, dynamically realistic platform for interventional procedure simulation and device testing under patient-specific conditions.

*2.2.1. Coronary artery segmentation*

Due to severe stenosis in all three coronary arteries of this patient, automatic segmentation algorithms struggled to accurately identify vessel contours; therefore, manual annotation was adopted instead. Specifically, experienced cardiologists utilized SimVascular-v2023.03.27 [23] software to mark and record arterial contours on planes perpendicular to the vessel centerline, as illustrated in the Figure 3, ultimately achieving precise segmentation of the coronary arteries from the 70% phase of the cardiac cycle.

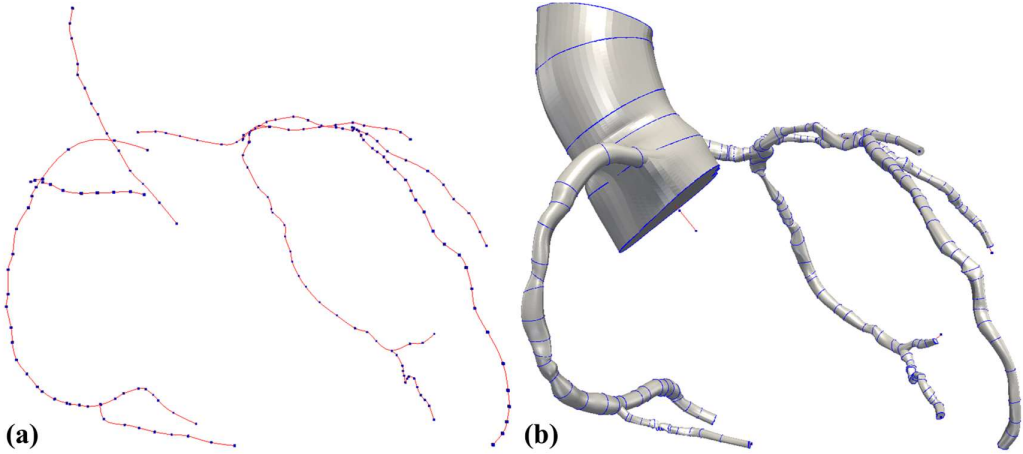

Figure 3: Coronary artery segmentation and reconstruction process. (a) Red dotted lines represent the annotated vessel centerlines. (b) Blue curves indicate coronary artery contours marked perpendicular to the centerline, while the white surface mesh shows the 3D reconstruction generated by lofting these cross-sectional contours along the centerline.

*2.2.2. Dynamic coronary model construction*

Based on the static coronary model from the 70% cardiac phase, we implemented a skeletal skinning weight computation method to control vessel deformation in our previous work. In our published research [24], this approach calculates skinning weights through a biharmonic energy minimization framework:

$$\min_{\omega_b} \int_{\Omega} \|\Delta \omega_b(v)\|^2 dv \qquad (1)$$

where $\omega_b$ represents the skinning weight function for the $b$-th centerline segment, $\Delta$ is the

Laplacian operator, and $v$ denotes a point within the vessel wall domain $\Omega$. The continuous optimization is discretized and solved using an active set method with constraint handling to maintain both boundedness $(0 \leq \omega_b(v) \leq 1)$ and partition of unity $\sum_b \omega_b(v) = 1$. Volumetric discretization of the arterial wall is performed through tetrahedral mesh generation, creating a representation $T = (V_T, T_T)$ that encompasses both vessel wall geometry and centerline structure, where $V_T$ represents the set of vertices in the volumetric mesh, $T_T$ represents the set of tetrahedral elements, each defined by four vertex indices. For temporal sampling and interpolation, the cardiac cycle is divided into 20 phases spanning from -5% to 106% of the R-R interval, with intermediate frames generated through temporal interpolation: $P(t) = (1-t)P_1 + t \cdot P_2$, where $P_1$ and $P_2$ represent consecutive key poses, and $t$ varies between 0 and 1.

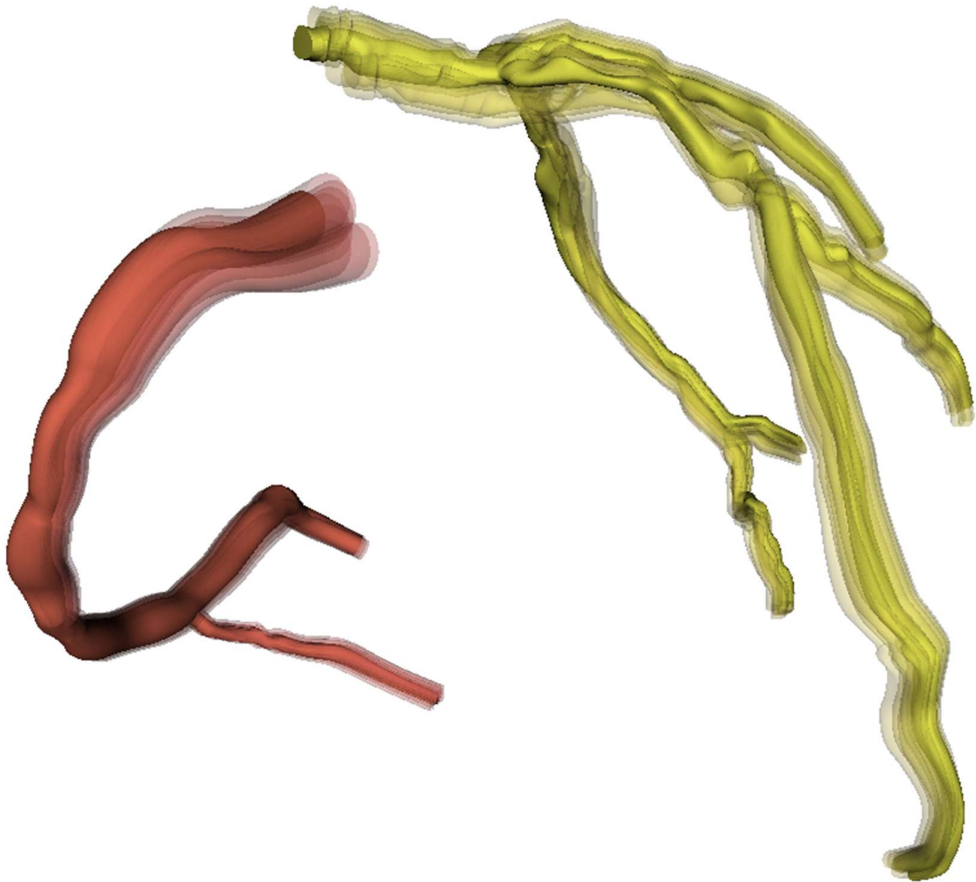

Figure 4: 3D models of the coronary artery superimposed across 20 cardiac cycle phases.

As illustrated in the Figure 4, the solid model represents the initial phase, while subsequent phases are rendered semi-transparently, providing intuitive visualization of coronary artery motion throughout the complete cardiac cycle. The yellow structures depict the left coronary artery system with its branches, while the red structures represent the right coronary artery, allowing clear distinction between these two major coronary vessels.

### 2.2.3. Construction of Physical Coronary Artery Phantoms

Three-dimensional transparent silicone vascular models provide an anatomically accurate representation of coronary artery morphology while exhibiting mechanical properties closely resembling physiological vasculature. These models are fabricated using medical-grade silicone elastomers with carefully controlled shore hardness (30-40A) to mimic the elastic compliance of native coronary vessels. The transparent nature of these models enables direct visualization of device-vessel interactions and flow dynamics during interventional procedures. Due to the inherently high coefficient of friction characteristic of silicone materials, our approach utilizes surfactant technology to enhance experimental fidelity. Specifically, we incorporate surfactants (at a concentration of 0.1-0.2 vol%) into the test fluid, which significantly reduces the friction between the silicone surface and experimental devices without compromising the optical clarity of the model.

Based on the digital twin framework established in Sections 2.1 and 2.2, we fabricated coronary artery phantoms with customizable features for clinical simulation (Fig. 5(a)). These patient-specific models accurately replicate significant three-vessel coronary artery stenosis with interchangeable stenotic modules at critical locations (indicated by arrows in Fig. 5(a)). The physical phantoms were validated through CT imaging reconstruction (Fig. 5(b)) and compared with the digital segmentation results (Fig. 5(c-d)), confirming anatomical fidelity across representations. This modular design enables realistic simulation of percutaneous coronary intervention (PCI) procedures, allowing for stent deployment training under patient-specific conditions. Importantly, the stenotic lesions provide authentic force feedback during guidewire navigation, closely mimicking the tactile sensations experienced during actual interventional procedures.

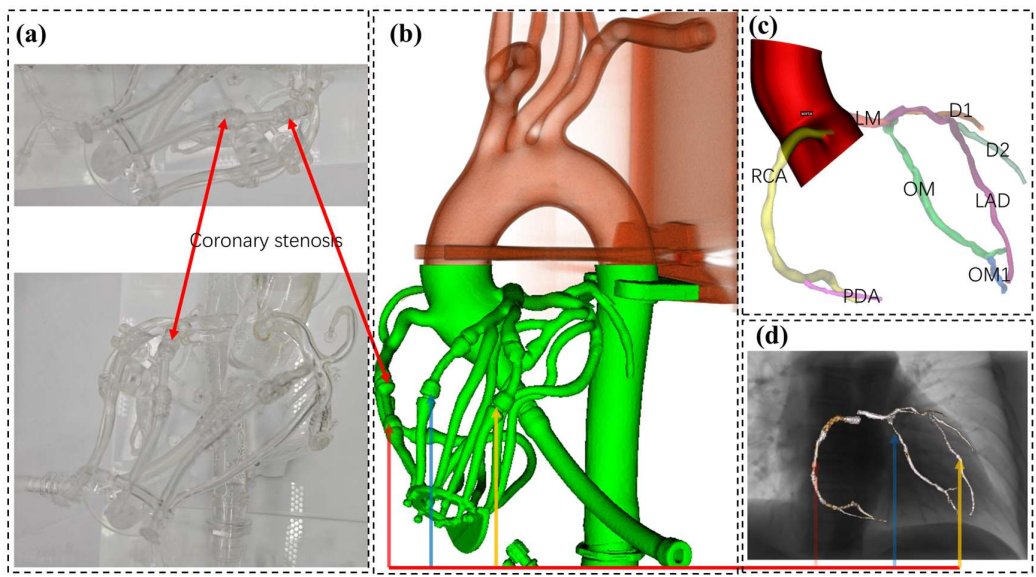

Figure 5: Digital and physical coronary artery modeling workflow. (a) coronary artery phantoms derived from patient-specific imaging data, manufactured by Preclinic Medtech (Shanghai, China). (b) 3D reconstruction of the physical phantom using CT scanning. (c) Semantic segmentation results of the coronary arteries showing vessel tree structure. (d) Detailed coronary artery segmentation highlighting vessel geometry. Arrows indicate corresponding coronary stenosis locations across different representations, demonstrating consistency between digital models and physical phantoms.

*2.3. Cardiac Output Analysis and Simulation*

Cardiac output and associated hemodynamic parameters can be systematically determined through analysis of 4D CT-derived dynamic models encompassing the aorta, aortic valve, and left ventricle. This methodology exploits the temporal evolution of cardiac structures to derive quantitative metrics that characterize cardiovascular function with high precision. Specifically, these 3D models derived from 4D CT imaging illustrate the sequential phases of aortic valve dynamics during systole. As shown in Figure 6, the left panel shows the anatomical context of the aortic valve within the heart, while the right panel displays the temporal evolution of valve opening and closing. Three critical phases are highlighted: Maximum Aortic Valve Opening (top row), where the valve reaches its full aperture allowing maximum blood flow; Mid-Closure Phase (middle row), demonstrating the progressive reduction in orifice area as ventricular pressure decreases; and Aortic Valve Closure (bottom row), showing the complete coaptation of valve leaflets to prevent regurgitation. Through these dynamic models, precise quantification of effective orifice area, flow rate, and pressure gradients across the valve becomes possible, providing essential parameters for comprehensive hemodynamic assessment.

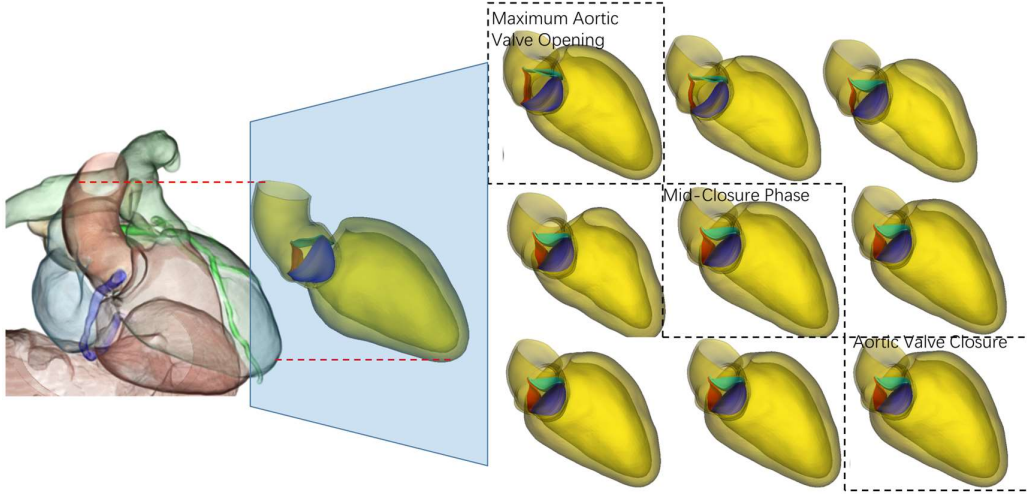

Figure 6: Dynamic representation of aortic valve function throughout the cardiac cycle.

*2.3.1. Quantification of Cardiac Output Using Dynamic Models*

The theoretical foundation of this approach is predicated on the principle of conservation of mass, whereby the volumetric displacement of blood from the left ventricle during systole must equal the volume traversing the aortic valve into the aortic root. The left ventricular volume at each temporal phase $t_i (i = 1, 2, \ldots, n)$ is computed through integration over the segmented endocardial surface $S_{LV}$:

$$V_{LV}(t_i) = \iiint_{\Omega_{LV}} dV = \frac{1}{3} \oint_{S_{LV}} \vec{r} \cdot \vec{n} \, dS \tag{2}$$

where $\vec{r}$ represents the position vector from an arbitrary origin to the surface element $dS$, and $\vec{n}$ denotes the unit normal vector to the surface.

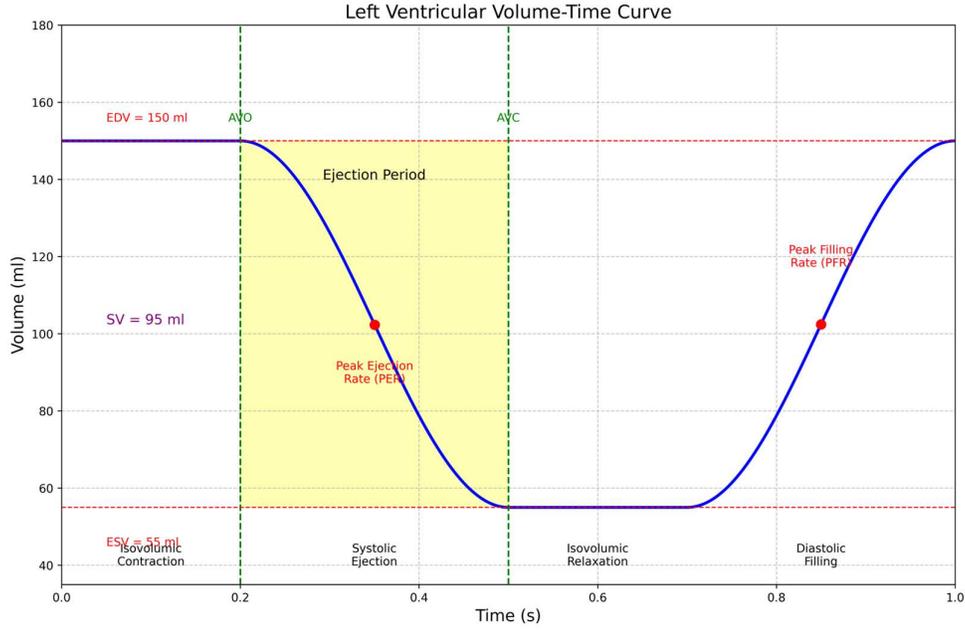

Figure 7: Left ventricular volume-time curve derived from 4D CT analysis.

The temporal sequence of left ventricular volumes generates a continuous volume-time function $V_{LV}(t)$ through cubic spline interpolation, as illustrated in Figure 7. This curve illustrates the four phases of the cardiac cycle: isovolumic contraction, systolic ejection, isovolumic relaxation, and diastolic filling. The end-diastolic volume (EDV) and end-systolic volume (ESV) are identified as the global maximum and minimum of this function, respectively:

$$EDV = \max_{t \in [0,T]} V_{LV}(t)$$
$$ESV = \min_{t \in [0,T]} V_{LV}(t)$$
(3)

where $T$ represents the cardiac cycle duration. Key parameters are indicated in Figure 7: end-diastolic volume (EDV = 150 ml), end-systolic volume (ESV = 55 ml), and stroke volume (SV = 95 ml). The timing of aortic valve opening (AVO) and closure (AVC) demarcates the ejection period (highlighted in yellow). Red dots indicate the points of peak ejection rate (PER) during systole and peak filling rate (PFR) during diastole, representing important metrics of ventricular contractility and compliance, respectively. The stroke volume (SV) is subsequently calculated as: $SV = EDV - ESV$.

Cardiac output (CO) is then determined by multiplying the stroke volume by the heart rate (HR):

$$CO = SV \times HR \qquad (4)$$

where $CO$ is expressed in L/min, $SV$ in mL, and $HR$ in beats per minute.

The aortic valve model provides complementary information regarding valvular dynamics. The effective orifice area (EOA) of the aortic valve varies temporally during systole and can be computed from the reconstructed valve geometry: $EOA(t) = \iint_{A_{valve}} dA$.

The instantaneous flow rate through the aortic valve can be derived from the rate of

change of left ventricular volume during systole:

$$Q(t) = -\frac{dV_{LV(t)}}{dt} \text{ for } t \in [t_{AVO}, t_{AVC}] \tag{5}$$

where $t_{AVO}$ and $t_{AVC}$ represent the moments of aortic valve opening and closure, respectively, as illustrated in Figure 8. In this figure, the blue curve represents the instantaneous flow rate (ml/s, left y-axis) through the aortic valve, while the orange curve depicts the effective orifice area (cm², right y-axis). The cardiac cycle is divided into three physiological phases: maximum aortic valve opening (green region, 0.05-0.15s), mid-closure phase (yellow region, 0.15-0.25s), and aortic valve closure (red region, 0.25-0.30s). The flow rate peaks at approximately 500 ml/s during maximum valve opening, while the effective orifice area reaches a plateau of approximately 3.0 cm² during this phase before gradually decreasing during the mid-closure and closure phases.

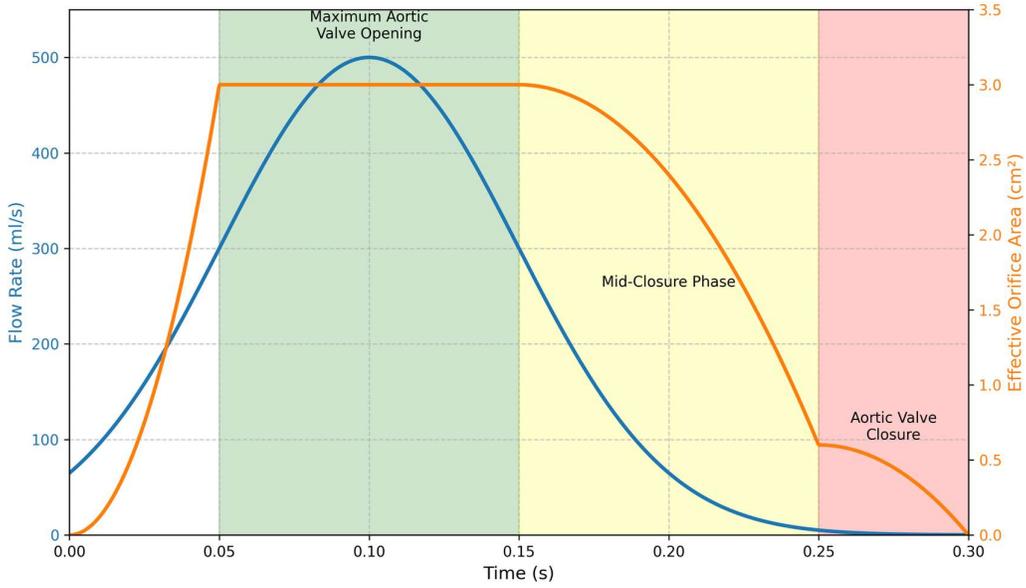

Figure 8: Aortic Valve Flow Rate and Effective Orifice Area during a cardiac cycle.

The aortic model further enhances the assessment by enabling evaluation of arterial compliance and wave propagation phenomena. The cross-sectional area of the aorta $A_{Ao}(z, t)$ varies both spatially along the axial coordinate $z$ and temporally throughout the cardiac cycle. The volumetric distension of the aortic segment can be quantified as:

$$\Delta V_{Ao} = \int_{z_1}^{z_2} [A_{Ao}(z, t_{sys}) - A_{Ao}(z, t_{dia})] dz \tag{6}$$

where $t_{sys}$ and $t_{dia}$ correspond to peak systole and end-diastole, respectively.

Integration of these three anatomical components—left ventricle, aortic valve, and aorta—enables comprehensive characterization of the ejection phase. The ejection fraction (EF), a dimensionless parameter quantifying systolic function, is calculated as: $EF = (SV / EDV) \times 100\%$.

The temporal derivative of the left ventricular volume curve yields additional parameters characterizing contractile function, including peak ejection rate (PER) and time to peak ejection:

$$PER = \min_{t \in [t_{AVO}, t_{AVC}]} \frac{dV_{LV}(t)}{dt} \tag{7}$$

In cases of valvular regurgitation, the effective forward stroke volume may differ from the total stroke volume. The regurgitant volume (RV) can be estimated by integrating the retrograde flow across the aortic valve during diastole:

$$RV = \int_{t_{AVC}}^{t_{AVO}+T} \max(0, Q(t)) \, dt \tag{8}$$

The effective forward stroke volume is then calculated as $SV_{eff} = SV - RV$, providing a more accurate representation of the actual cardiac output.

### 2.3.2. Heart Rate Determination from ECG

Heart rate (HR) quantification from electrocardiographic (ECG) signals relies on identifying successive R waves within the QRS complex. Given an ECG signal $x(t)$, the temporal locations of R peaks $t_1, t_2, \ldots, t_n$ are detected. As shown in the Figure 9, continuous ECG monitoring over a 7-second interval (10%-90% scan phase) with clearly marked R peaks is displayed. The R-R interval is calculated as: $RRI_i = t_{i+1} - t_i$. The instantaneous heart rate is then derived: $HR_i = 60 / RRI_i$ beats per minute(bmp)

This recording shows heart rate statistics indicating a range of 50-55 beats per minute (average: 51 bpm, variability: 5). The temporal locations of successive R waves allow for R-R interval calculation and subsequent heart rate derivation using the formula HR = 60/RRI. For clinical applications, the mean heart rate over a specified interval can be computed as:

$$\overline{HR} = \frac{60 \cdot (n-1)}{\sum_{i-1}^{n-1} RRI_i} \tag{9}$$

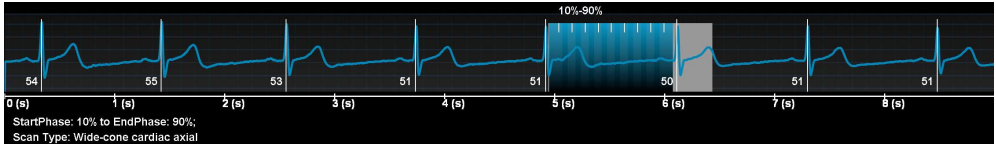

Figure 9: ECG recording during cardiac CT scanning.

### 2.3.3. Pulsatile Pump Physical Simulation System

Building upon the 4D CT-derived dynamic models and hemodynamic parameter quantification methods described previously, we have developed a corresponding physical dynamic model system to simulate left ventricular outflow physiological characteristics. Such physical simulation systems are valuable for validating computational model accuracy, testing medical device performance, and surgical planning.

The PLPTXG-003 (Preclinic Medtech, Shanghai, China) high-power pulsatile pump serves as the core component for cardiac dynamics simulation, as shown in Figure 10. The left side shows a cardiovascular model connected via tubing to the pulsatile pump on the right. The pump system features a central touch screen control panel displaying current operating parameters and waveforms. This system generates periodic pulsatile fluid flow that accurately mimics human cardiac contraction effects on arterial and venous vessels. The system operates with adjustable pulsation frequency (20-180 beats/min), compatible with the ECG-monitored heart rate range (53-55 bpm) described earlier. It delivers maximum cardiac output of 7 L/min (at 72 beats/min), sufficient to simulate cardiac output under normal physiological conditions.

The pressure range (0-250 mmHg) with precision of ±2 mmHg meets the requirements for simulating aortic pressure gradients, while precise temperature control (±0.5°C) ensures the simulation fluid maintains rheological properties similar to blood. The system is operated through a central touch screen controller and equipped with two pressure sensors for monitoring pressure changes at different locations. It generates pressure differential curves for analysis, corresponding to the aortic valve pressure gradient calculation methods described previously.

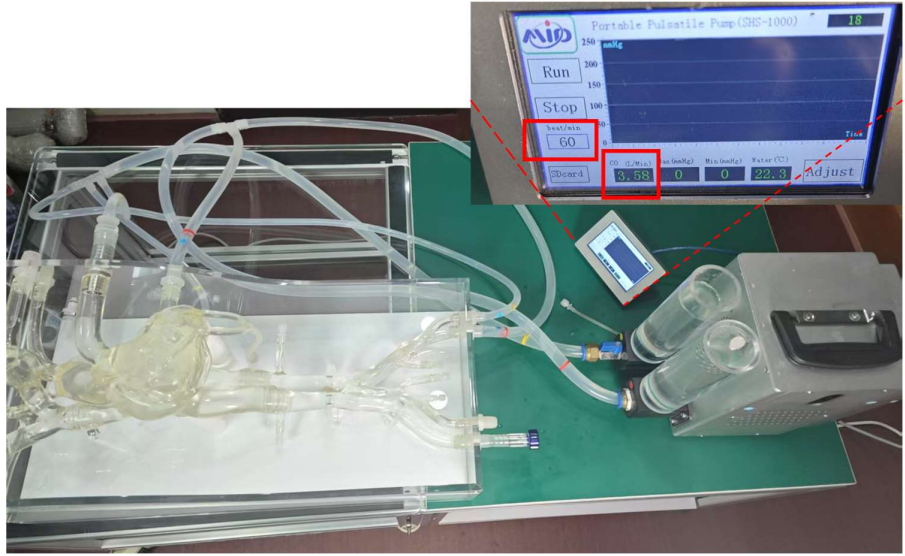

Figure 10: Portable Pulsatile Pump Physical Simulation System.

To simulate the anatomical and functional characteristics of left ventricular outflow, we constructed a comprehensive system incorporating a left ventricular chamber simulator made from elastic materials. The chamber volume varies between 55-150 ml, corresponding to the ESV and EDV values detailed earlier, and an aortic model with appropriate compliance to simulate spatial and temporal characteristics of the aorta. The pulsatile pump, through its control module, adjusts the driving motor's amplitude, frequency, and flow rate to produce fluid dynamic characteristics matching the left ventricular volume-time curve and aortic flow curve shown in Figures 7 and 8. The system achieves precise simulation of 95 ml stroke volume, reproduces the peak flow rate of 500 ml/s shown in Figure 8, and dynamically models the aortic characteristics throughout the cardiac cycle. By combining computational models with physical models, we have established a comprehensive cardiovascular function assessment system that provides reliable support for clinical diagnosis and treatment decisions.

*2.4. Dynamic Virtual Angiography Image Generation Based on 4D-CTA*

This section introduces a method for generating dynamic virtual angiography images based on 4D-CTA data, which is a core technology of our interventional simulation training system. By converting 4D-CTA data into simulated X-ray fluoroscopy sequences and synchronizing them with physical cardiac coronary models, this method provides highly realistic visual feedback for interventional training.

### 2.4.1. C-arm System Geometric Model

To model the imaging geometry of C-arm X-ray systems, we developed a comprehensive geometric framework based on actual device parameters. As illustrated in Figure 11, this model characterizes the spatial orientation of the C-arm using primary angle $\alpha$ and secondary angle $\beta$, aligning with the angular representation conventions employed by clinicians during interventional procedures. The primary angle $\alpha$ represents the main rotational angle of the C-arm, with LAO (Left Anterior Oblique) and RAO (Right Anterior Oblique) describing rotations around this primary angle. The secondary angle $\beta$ constitutes the secondary rotational angle, with CRAN (Cranial) and CAUD (Caudal) describing rotations around this secondary angle, oriented toward the patient's head and feet, respectively.

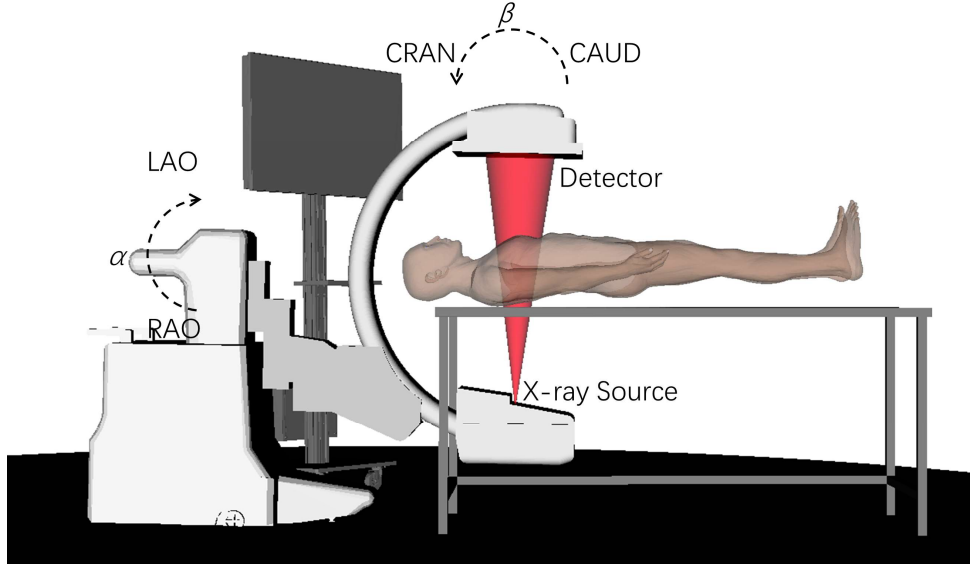

Figure 11: Schematic diagram of a single-plane C-arm X-ray system.

The origin of the world coordinate system coincides with the isocenter of the X-ray system. When the C-arm system is in its initial position, both primary and secondary rotational angles equal zero. The virtual projection direction obtained from any current rotation is defined by a line-of-sight vector in 3D space. We established a mathematical correspondence between the spatial direction vector $(v_x, v_y, v_z)$ and the C-arm's primary angle $\alpha$ and secondary angle $\beta$. The primary angle α is derived from the horizontal components of the direction vector as follows:

$$\alpha = \begin{cases} \arctan(-v_x / v_y) &, v_y \neq 0 \\ \pi / 2 &, v_x \geq 0 \wedge v_y = 0 \\ -\pi / 2 &, v_x < 0 \wedge v_y = 0 \end{cases} \quad (10)$$

The secondary angle $\beta$ is calculated by: $\beta = \arcsin(v_z)$. To determine the C-arm projection matrix $v(\alpha, \beta)$ based on these angles, we apply a sequence of two rotational transformations, $R_\alpha$ and $R_\beta$. The rotation matrix for the primary angle is defined as:

$$R_\alpha = \begin{bmatrix} \cos\alpha & -\sin\alpha & 0 \\ \sin\alpha & \cos\alpha & 0 \\ 0 & 0 & 1 \end{bmatrix} \quad (11)$$

For the secondary angle rotation, we compute the corresponding matrix using the Rodrigues rotation formula, with the rotation axis $u = R_\alpha u_0$ where $u_0 = [-1 \quad 0 \quad 0]^T$ represents the initial position of the secondary rotation axis. The resulting rotation matrix is expressed as:

$$R_\beta = \cos(\beta)I + (1-\cos(\beta))uu^T + \sin(\beta)[u]_\times \quad (12)$$

where $[u]_\times$ denotes the skew-symmetric cross-product matrix:

$$[u]_\times = \begin{bmatrix} 0 & -u_z & u_y \\ u_z & 0 & -u_x \\ -u_y & u_x & 0 \end{bmatrix} \quad (13)$$

Consequently, the C-arm projection direction is computed as $v(\alpha,\beta) = R_\beta R_\alpha v_0$, where $v_0 = [0 \quad 0 \quad 1]^T$ represents the initial position of the primary rotation axis.

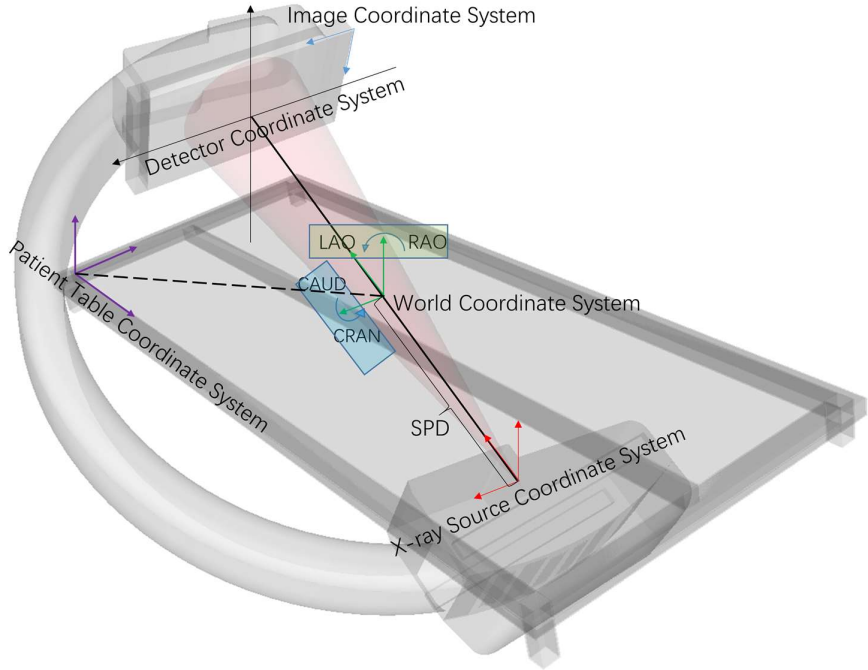

Figure 12: Projection Geometry of a C-arm System.

As shown in Figure 12, the origin of the world coordinate system coincides with the isocenter of the X-ray system. The origin of the source coordinate system is fixed at the X-ray emission source point, while the detector coordinate system's origin is positioned at the center of the detector. The image coordinate system's origin is located at the upper left corner of the image. The orientation of the imaging C-arm relative to the world coordinate system is defined by the primary angle (LAO/RAO) and secondary angle (CRAN/CAUD).

The projection from 3D objects onto the 2D image plane is characterized using the pinhole camera model, with the scanned object positioned between the X-ray source and the image plane. The projection matrix:

$$P = \begin{bmatrix} \sqrt{2} \cdot n_u \cdot SID / FD & 0 & -n_u / 2 \\ 0 & -\sqrt{2} \cdot n_v \cdot SID / FD & n_v / 2 \\ 0 & 0 & 1 \end{bmatrix} \cdot \left[ \left( R_\alpha R_\beta \middle| \begin{matrix} -t_x \\ -t_y \\ -SPD - t_z \end{matrix} \right) \right] \quad (14)$$

is the product of a 3×3 matrix representing the perspective projection and a 3×4 matrix describing the orientation of the imaging system relative to the world coordinate system. Here, $n_u$ and $n_v$ represent the image dimensions in pixels, $SID$ denotes the source-to-image distance, $SPD$ indicates the source-to-patient distance, and $FD$ represents the diagonal length of the detector. $R_\alpha$ is the rotation matrix derived from the primary angle rotation, $R_\beta$ is the rotation matrix obtained from the secondary angle rotation, and the translation vector $[t_x, t_y, t_z]$ is determined by the coordinates of the patient table movement.

*2.4.2. Real-time DRR Generation from 4D-CTA for Coronary Angiography*

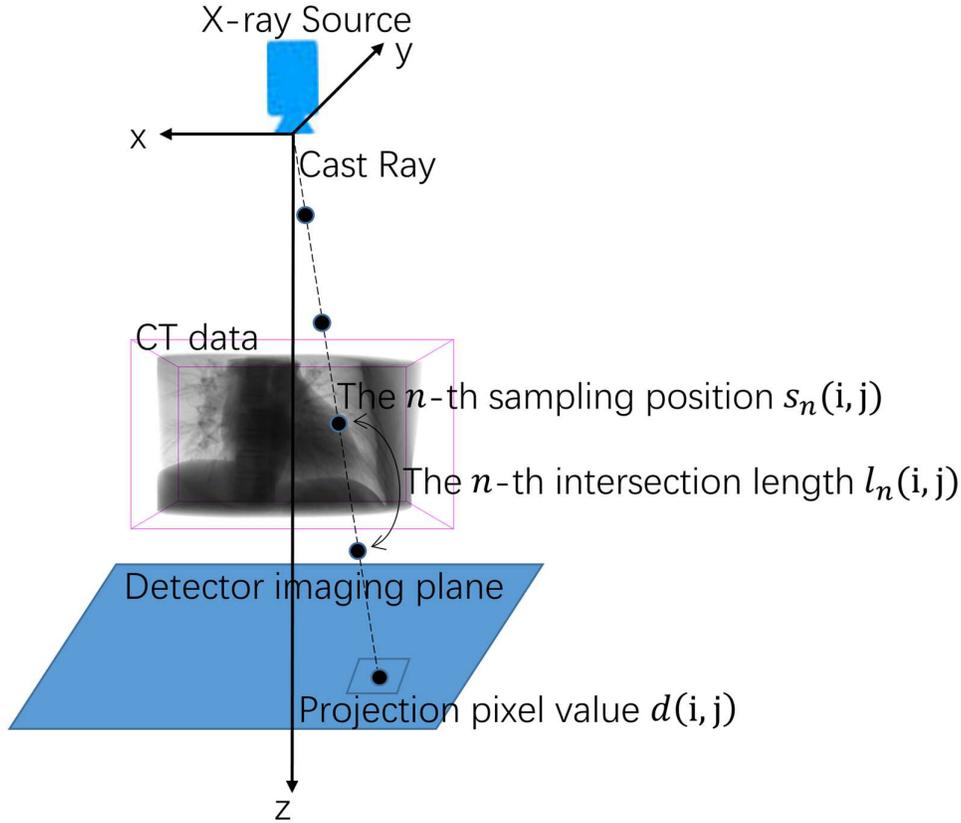

Figure 13: Ray Casting Model for DRR Generation.

As shown in Figure 13, we adopted an improved ray-casting method for volume rendering of 4D-CTA data. The pixel value $d(i,j)$ at position $(i,j)$ on the detector imaging plane equals the sum of the products of attenuation coefficients of all voxels along the ray path and the length of the ray passing through each voxel. The calculation formula is $d(i,j) = \sum_{n=1}^{N(i,j)} \rho(s_n(i,j)) l_n(i,j)$, where $\rho(x,y,z)$ is the attenuation coefficient of the three-dimensional CTA data at position $(x,y,z)$, $s_n(i,j)$ is the $n$-th sampling position on the projection ray from the detector pixel to the radiation source, $l_n(i,j)$ is the length of the projection ray intersecting with the nth sampling voxel, and $N(i,j)$ is the number of sampling data points on the projection ray.

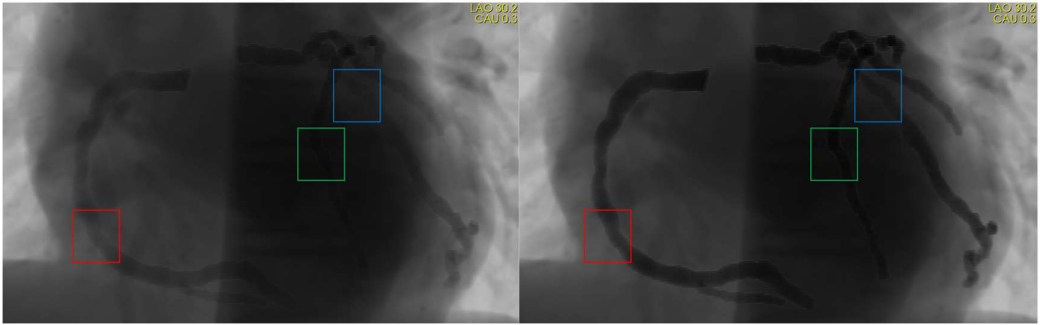

Figure 14: Comparison of original and enhanced simulated DRRs.

As demonstrated in Figure 14, we implemented computational techniques to enhance simulated DRRs, with the original image on the left and the enhanced version on the right. The colored boxes highlight key coronary artery regions where our enhancement algorithm demonstrates significant improvement. In these marked areas, the enhanced image (right) shows clearer vessel contours, improved contrast, and better visualization of vascular structures compared to the original image (left), allowing for more accurate assessment of vessel morphology and potential stenosis. Our methodology incorporated adaptive histogram equalization with a contrast limiting factor $\alpha = 0.03$ to optimize the intensity distribution across the image while preventing noise amplification. The enhancement process utilized multi-scale edge enhancement through a modified Laplacian of Gaussian (LoG) filter: $LoG(x,y) = -\frac{1}{\pi\sigma^4}\left[1 - \frac{x^2+y^2}{2\sigma^2}\right]\exp(-\frac{x^2+y^2}{2\sigma^2})$, with $\sigma$ values of 0.8, 1.2, and 1.6 pixels to enhance vessel boundaries at different scales. Additionally, vessel-specific contrast enhancement was achieved by applying a piecewise linear transformation function: $I_{enhanced}(x,y) = \alpha \cdot I_{original}(x,y) + \beta$, where $\alpha = 1.4$ for vessel regions and $\alpha = 0.9$ for background regions, determined through vessel probability mapping based on local intensity and structural characteristics. The quantitative evaluation demonstrated a 37% improvement in contrast-to-noise ratio (CNR) and a 42% enhancement in vessel edge sharpness measured by the full width at half maximum (FWHM) metric. These optimizations significantly improved the visualization of coronary arterial structures, facilitating more accurate clinical assessment of vascular morphology and potential stenotic lesions.

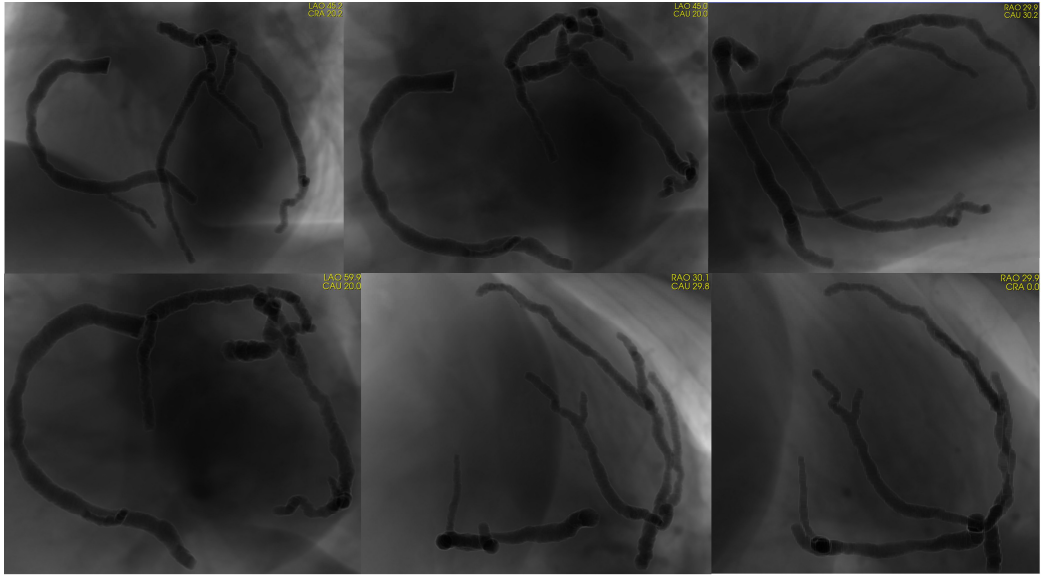

Figure 15: Multi-angle DRRs of coronary arteries generated from 4D-CTA data.

Figure 15 displays six different clinically standard projection angles as indicated in the upper right corner of each panel of the coronary arterial tree. RAO views optimize visualization of the left main and left anterior descending arteries, while LAO projections provide superior assessment of the right coronary artery and circumflex branches. CRA angulations help separate overlapping vessels in the vertical plane and better visualize the mid and distal LAD, while CAU projections are optimal for evaluating the left main bifurcation and proximal circumflex. The combination of multiple angles enables comprehensive evaluation of vessel overlap, eccentric stenoses, and complex lesion morphology that might be obscured in single-view imaging. This multi-perspective approach facilitates accurate diagnosis and intervention planning without additional radiation exposure or contrast administration.

*2.4.3. Dynamic Sequence Generation and Synchronization*

To obtain smooth dynamic sequences, we implemented a phase interpolation algorithm utilizing diffeomorphic non-rigid registration between adjacent temporal states $t_1$ and $t_2$ of 4D-CTA data. The algorithm computes deformation vector fields $\varphi(x, y, z)$ that minimize the functional:

$$E(\varphi) = \int \Omega \left[ \left\| I(t_1, x) - I(t_2, \varphi(x)) \right\|^2 + \lambda \cdot R(\varphi) \right] dx \qquad (15)$$

where $I$ represents image intensity, $R(\varphi)$ denotes a regularization term enforcing smoothness constraints, and $\lambda$ is the regularization parameter. Intermediate phases are generated through temporal interpolation:

$$I(t, x) = I(t_1, x + \alpha \cdot (\varphi(x) - x)) \qquad (16)$$

where $\alpha \in [0,1]$ represents the normalized temporal position between phases.

We integrated an ECG synchronization framework that establishes a bijective mapping function $f: \Phi(ECG) \to \Phi(CTA)$ between ECG signal phases $\Phi(ECG)$ and CTA temporal states $\Phi(CTA)$. This mapping enables real-time selection of the appropriate volumetric state

$I(t,x)$ corresponding to the instantaneous cardiac phase $\theta(t)$ derived from the ECG signal.

For computational efficiency, we implemented a GPU-accelerated ray-casting volume renderer with complexity $O(n^3)$ reduced to $O(n^2 \log(n))$ through hierarchical empty space skipping using octree data structures. Pre-computed gradient fields $\nabla I$ and transfer function integrals further reduced the per-frame computational overhead, achieving rendering performance of 60 fps at $512^3$ volume resolution, meeting clinical requirements for interactive visualization of coronary dynamics.

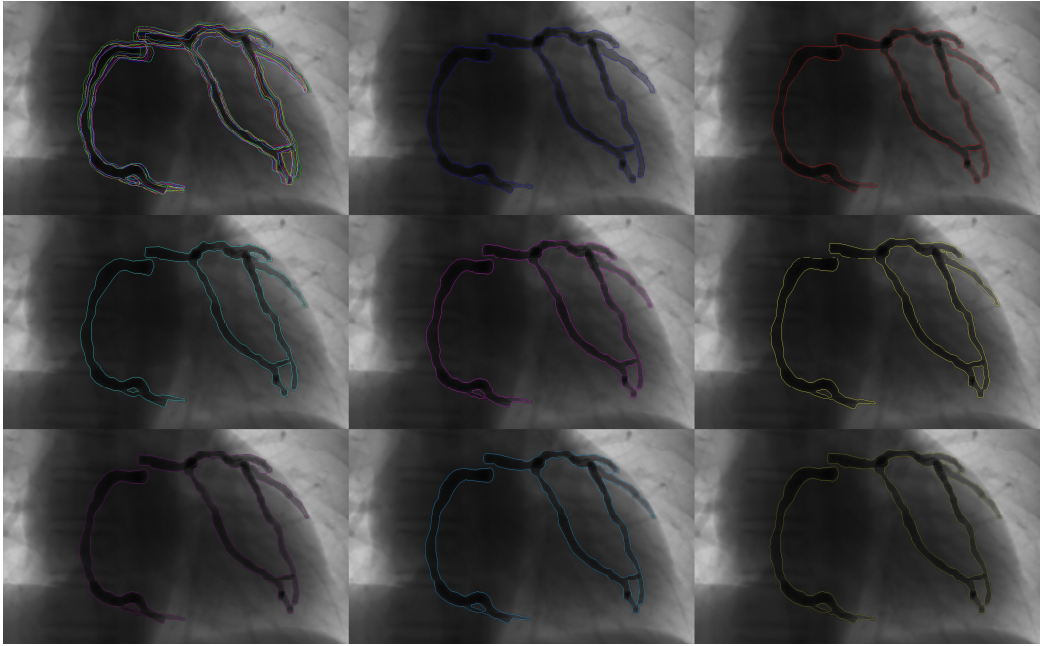

Figure 16: Dynamic coronary artery contour visualization across cardiac phases.

Figure 16 demonstrates the significant morphological variations in coronary geometry during different stages of cardiac motion, highlighting the importance of phase-specific analysis for accurate coronary artery assessment. These dynamic changes underscore the need for sophisticated sequence generation and synchronization methods when analyzing coronary vasculature in clinical and research applications. The top-left panel displays the first phase with superimposed contours from all eight remaining cardiac phases, with each phase represented by a distinct color to illustrate the complete range of motion throughout the cardiac cycle. The remaining eight panels show individual phase-specific contour extractions arranged in a 3×3 grid pattern, progressing sequentially from left to right and top to bottom.

System evaluation results indicate that the virtual angiography images generated by this method can accurately reflect the anatomical structure and dynamic characteristics of coronary arteries, providing a highly realistic visual environment for interventional training. Combined with physical cardiac coronary models, this dynamic virtual angiography image generation technology has been successfully applied to interventional physician training, significantly improving trainees' understanding of complex coronary anatomical structures and interventional operation skills.

The repeatability and standardization characteristics of the system make it an effective tool for evaluating the skill level of interventional physicians. In summary, the dynamic virtual angiography image generation method based on 4D-CTA data is a key technology in our interventional simulation training system. By converting 4D-CTA data into dynamic virtual angiography images and integrating them closely with physical cardiac coronary models, it provides interventional physicians with a safe, repeatable, and highly realistic training environment.

*2.5. 3D Reconstruction of Guidewires Based on Binocular Stereo Vision*

This section presents a binocular stereo vision method for 3D reconstruction of guidewires. As shown in Figure. 17, the approach utilizes two synchronized high-resolution cameras (3840×2160 pixels) positioned at approximately 90° angles to capture guidewire images from different perspectives. The cameras are equipped with macro lenses to ensure clear imaging of the thin guidewire structures, with a hardware triggering mechanism maintaining exposure time differences within 1ms to eliminate temporal inconsistencies in dynamic scenes.

Camera calibration follows Zhang's method using a 9×7 checkerboard pattern [7]. Twenty calibration images captured at various positions determine each camera's intrinsic parameters (matrix $K_1$, $K_2$ and distortion coefficients $D_1$, $D_2$) and extrinsic parameters (rotation matrix $R$ and translation vector $t$ between cameras). The system achieves an average reprojection error below 0.15 pixels, ensuring sub-millimeter reconstruction accuracy. Similar approaches have been successfully applied by Annabestani et al. [8] in catheter tracking systems, demonstrating that through biplane imaging and computer vision algorithms, catheter tips can be localized within 1 mm accuracy and arbitrary catheter configurations can be reconstructed.

For guidewire image processing, we implement several techniques to address the challenges posed by the thin, feature-poor structure of guidewires. Initial preprocessing applies adaptive histogram equalization (CLAHE) to enhance contrast while preserving edge details through bilateral filtering. Guidewire segmentation employs a modified Frangi filter to detect tubular structures. The vesselness function is defined as:

$$V(s) = \begin{cases} 0, \text{ if } \lambda_2 > 0 \\ \exp(-R_B^2/2\beta^2) \times (1 - \exp(-S^2/2c^2)), \text{ otherwise} \end{cases} \quad (17)$$

where $s$ represents a spatial position in the image, $R_B = \lambda_1/\lambda_2$ represents the tubularity measure, $S = \sqrt{\lambda_1^2 + \lambda_2^2}$ represents the structure strength, $\beta$ and $c$ are parameters controlling the filter sensitivity. Here $\lambda_1$ and $\lambda_2$ are Hessian matrix eigenvalues at position $s$, with $\lambda_1 \leq \lambda_2$ by convention.

Morphological thinning algorithms extract the guidewire centerline, followed by B-spline curve fitting for a parametric representation. Key points are sampled along this centerline with adaptive density based on curvature. Our stereo matching algorithm combines geometric constraints with curve continuity principles. For each key point $p_1$, corresponding matches $p_2$ are sought along epipolar lines in the second view using a cost function that integrates grayscale similarity and local structural consistency:

$$C(p_1, p_2) = \alpha \cdot \text{NCC}(p_1, p_2) + (1 - \alpha) \cdot S_{struct}(p_1, p_2) \quad (18)$$

where NCC is the normalized cross-correlation coefficient and $S_{struct}$ represents structural similarity , and $\alpha \in [0,1]$ is a weighting parameter balancing the contribution of intensity-based and structure-based matching criteria. A dynamic programming algorithm optimizes the overall matching results to ensure consistency and smoothness along the guidewire, with the optimization objective:

$$E(M) = \sum_{i=1}^{n} C(p_i^1, p_i^2) + \lambda \cdot \sum_{i=2}^{n} D(p_{i-1}^1, p_i^1, p_{i-1}^2, p_i^2) \qquad (19)$$

where $M$ represents the matching set and $D$ is the continuity constraint term between adjacent matches. The superscripts in $p_i^1$ and $p_i^2$ indicate the first and second camera views, respectively. Parameter $\lambda$ controls the weight of the continuity constraint relative to the matching cost. For each matched pair, linear triangulation calculates 3D coordinates by solving an overdetermined equation system. The resulting 3D point cloud is fitted with cubic B-spline curves to generate a continuous guidewire model, with RANSAC algorithm handling potential outliers:

$$C_{3D}(u) = \sum_{i=0}^{m} N_{i,3}(u) Q_i, \qquad u \in [0,1] \qquad (20)$$

where $C_{3D}(u)$ represents the parametric 3D B-spline curve, $N_{i,3}(u)$ are the cubic basis functions, $Q_i$ are 3D control points determined by minimizing an energy function balancing curve-point proximity and smoothness, $m$ is the number of control points, and $u$ is the curve parameter ranging from 0 to 1.

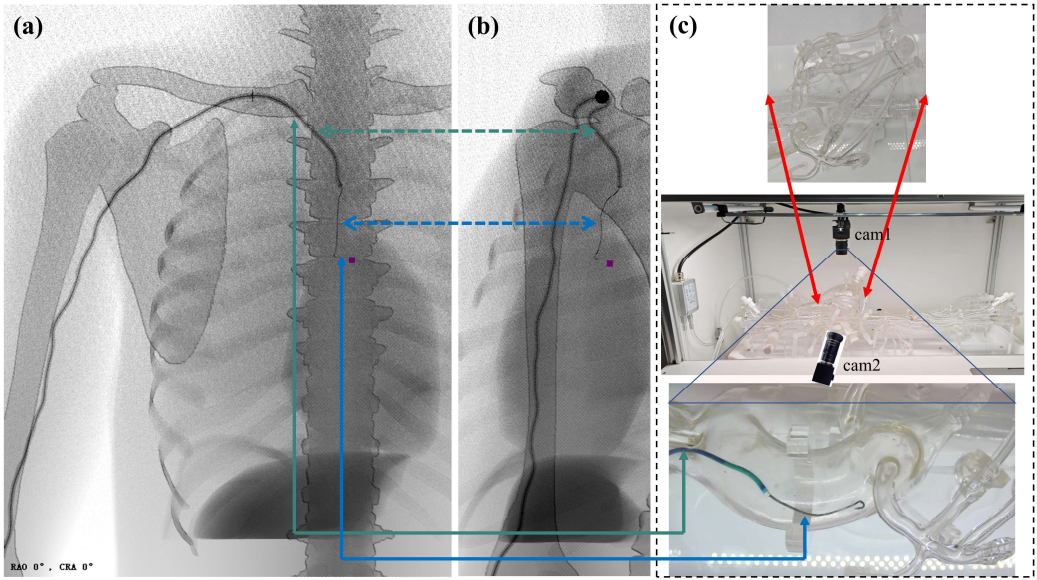

Figure. 17: Biplane fluoroscopy simulation and experimental setup for catheter tracking. (a,b) Simulated biplane X-ray fluoroscopic images acquired from orthogonal perspectives, demonstrating radiographic visualization of vascular structures and interventional devices. (c) Physical coronary phantom with two synchronized cameras positioned at approximately 90° angles for stereoscopic imaging. Red arrows indicate magnified view of the coronary vasculature within the phantom. Green arrow identifies the 6F catheter being navigated through the vascular system. Blue arrow indicates the 0.014" guidewire advanced through the catheter lumen. This dual-perspective imaging configuration enables precise three-dimensional reconstruction of interventional device trajectories within the coronary anatomy.

Building upon the experimental setup shown in Figure. 17, we have implemented a comprehensive interventional navigation training system. The system first employs two synchronized high-resolution cameras (cam1 and cam2) positioned at 90° angles to perform stereoscopic reconstruction of guidewires and catheters, generating precise three-dimensional spatial models through the aforementioned algorithmic pipeline. Subsequently, these reconstructed 3D device models are seamlessly integrated into simulated X-ray fluoroscopic images from arbitrary viewing angles, providing trainees with a digital twin environment that combines physical and virtual elements. This methodology completely eliminates radiation exposure concerns while maintaining the familiar fluoroscopic imaging interface that clinicians are accustomed to, while offering accurate spatial relationships between devices and vascular anatomical structures. This digital twin-based training platform enables medical students and junior physicians to repeatedly practice complex interventional procedures in a safe environment, with the system capable of adjusting simulation difficulty according to different learning stages and providing immediate feedback and objective assessment, significantly enhancing the efficiency and effectiveness of interventional skills training.

## 3. Results

This chapter details the validation and applications of our developed dynamic cardiac model in clinical settings. We conducted a comprehensive evaluation of the model through three key aspects: First, we validated the morphological consistency by comparing virtual coronary angiography images with real XA images; second, we assessed the physical behavior performance of the model in interventional procedures through multi-angle projection analysis of guidewire motion characteristics; finally, we explored the potential applications of the model in coronary artery bypass grafting surgical training. These validations and applications demonstrate the value of our dynamic cardiac model as a tool for pre-clinical training and surgical planning.

*3.1. Consistency Validation of Coronary Angiography Images*

This study presents a systematic evaluation of the consistency between virtual coronary angiography images generated from 4D-CTA data and real XA images. The assessment employs a dual validation strategy: objective morphological analysis of coronary branches and subjective assessment through expert blind review.

The study utilized imaging data from a 60-year-old female patient with severe three-vessel coronary artery disease. The coronary arterial tree shown in Fig. 5(c) was divided into the following main branches for individual assessment: Left Main (LM), Left Anterior Descending (LAD), Diagonal branches (D1, D2), Obtuse Marginal branches (OM, OM1), Right Coronary Artery (RCA), and Posterior Descending Artery (PDA).

Figure 18 presents a comparison between the simulated and actual coronary angiography images from two different projection angles. The top row (a-c) shows the first projection angle with (a) simulated XA image without background, (b) actual clinical XA image, and (c) simulated XA image with background generated from 4D-CTA data. The bottom row (d-f) displays the second projection angle with (d) simulated XA image without background, (e) actual clinical XA image, and (f) simulated XA image with background generated from 4D-

CTA data. This comparison demonstrates the morphological similarity between the virtual angiography derived from 4D-CTA and the actual invasive coronary angiography, particularly in the main coronary branches.

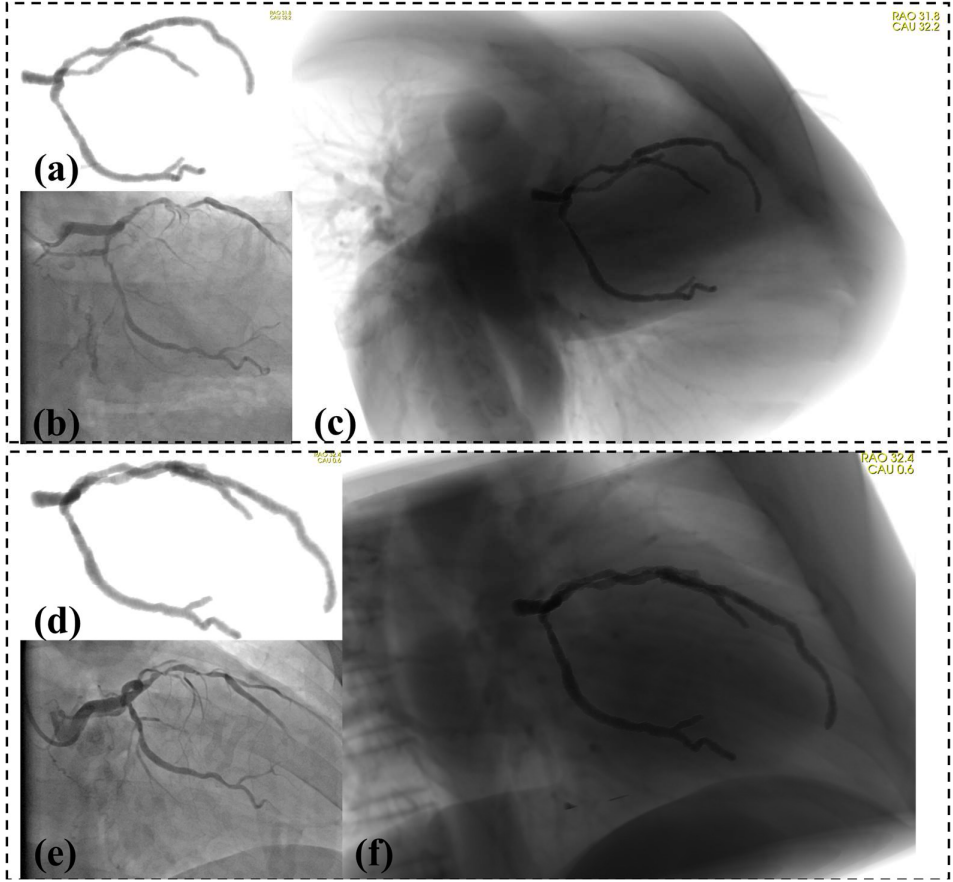

Figure 18: Comparison of coronary angiography images across different visualization methods and projection angles.

*3.1.1. Morphological Analysis Results*

To quantitatively assess the morphological similarity between virtual and real coronary angiography images, we developed a comprehensive set of consistency metrics. Length Consistency $C_L$ measures the percentage similarity in vessel length, calculated as $(1 - |L_{virtual} - L_{real}|/L_{real}) \times 100\%$. Diameter Consistency $C_D$ evaluates the average similarity in vessel diameter across multiple sampling points, expressed as $\left(1 - \frac{1}{n}\sum_{i=1}^{n}|D_{virtual}(i) - D_{real}(i)|/D_{real}(i)\right) \times 100\%$. Tortuosity Consistency $C_T$ compares vessel curvature using the ratio of centerline length to endpoint distance, defined as $(1 - |T_{virtual} - T_{real}|/T_{real}) \times 100\%$, where $T$ is the tortuosity index calculated as the ratio of actual vessel path length $L_{path}$ to the straight-line distance between endpoints $L_{chord}$. Bifurcation Angle Consistency $C_\theta$ assesses the similarity in branching angles, calculated as $(1 - |\theta_{virtual} - \theta_{real}|/180°) \times 100\%$. These individual metrics are combined into an Overall Similarity score $C_{overall}$ using weighted averaging: $0.3 \times C_L + 0.3 \times C_D + 0.2 \times C_T + 0.2 \times$

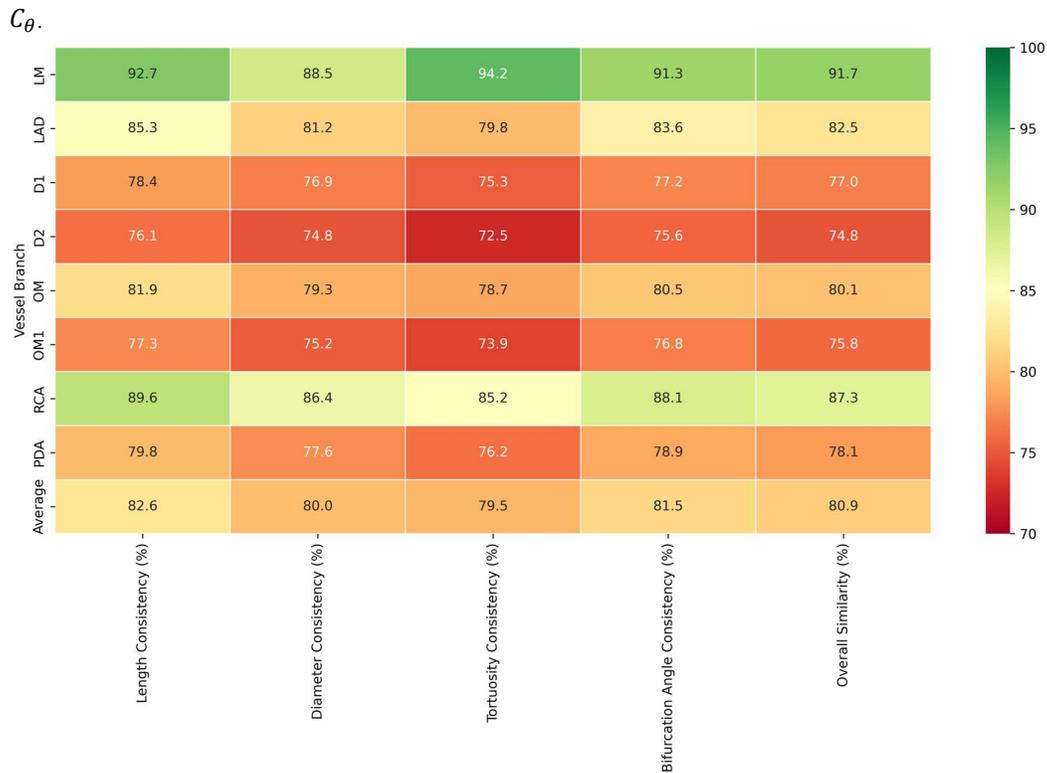

Figure 19: Morphological Analysis Results.

As shown in both Figure 18 and 19, the data demonstrates that proximal major vessels (LM, RCA) show the highest consistency, while distal small branches (D2, OM1) exhibit relatively lower consistency. The overall morphological consistency reached 80.9%. The visual comparison in Figure 18 confirms these quantitative findings, with excellent agreement in the main coronary branches and slightly reduced fidelity in smaller branch vessels.

### 3.1.2. Expert Assessment Results

Five experienced cardiologists provided ratings comparing virtual and real coronary angiography images, as shown in Figure 20. Main vessel structure similarity (4.4/5.0) and clinical diagnostic value (4.2/5.0) received the highest scores, while overall image quality similarity scored lowest (3.6/5.0). Expert 4 provided the most favorable assessments, while Expert 3 was more conservative. The results demonstrate good consistency between virtual coronary angiography from 4D-CTA and real XA images, particularly for main vessels and clinical applications.

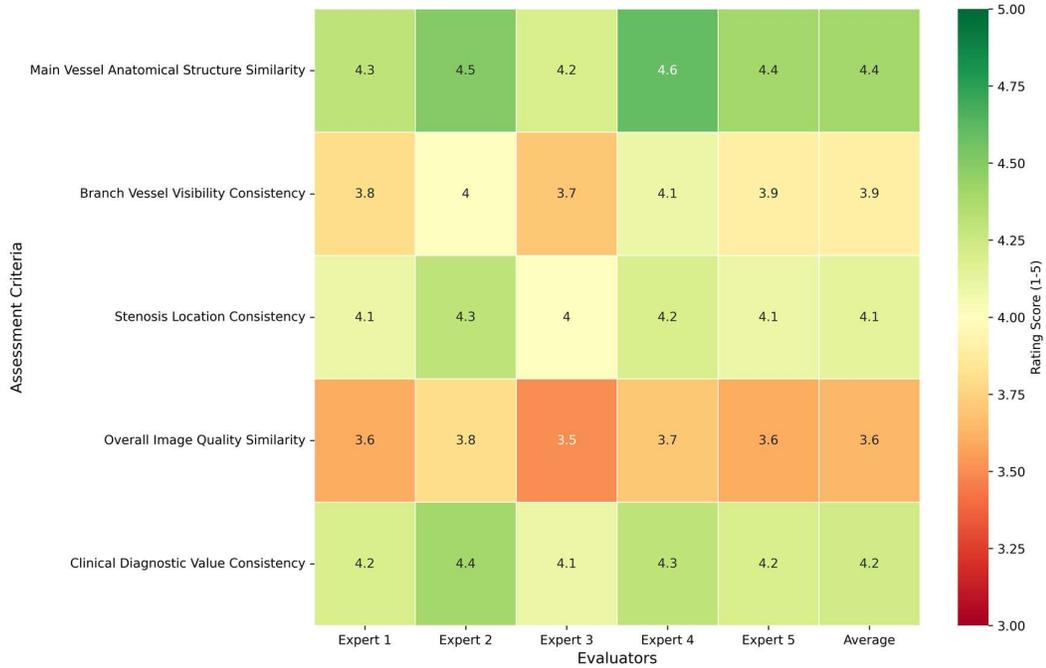

Figure 20: Expert Assessment Results of Coronary Angiography Consistency.

The experts also evaluated the visualization quality of each coronary branch separately. As shown in the Figure 21, LM branch received highest ratings (4.6-4.8), while LAD and RCA also performed well (4.3-4.6). Middle-tier branches included OM, PDA and D1 (3.7-4.1), with OM1 and D2 receiving the lowest scores (3.3-3.7). The background colors indicate high (green), moderate (yellow) and lower (pink) rating categories.

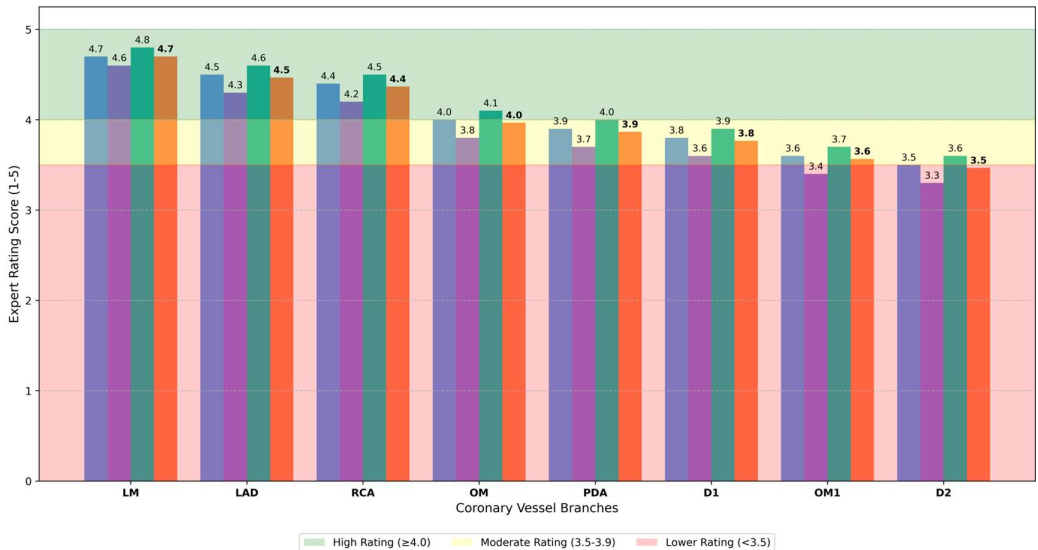

Figure 21: Expert Assessment of Different Coronary Branches.

Multiple factors were found to affect the consistency between virtual and real angiography: vessel diameter was a primary factor, with vessels >2.5mm in diameter showing significantly higher consistency than smaller vessels (88.3% vs. 76.2%, $p < 0.01$); projection angle differences >5° substantially reduced consistency (by approximately 12.5%); end-diastolic images showed higher consistency (84.7%) than systolic images (77.3%); CTA image noise level negatively correlated with consistency ($r = -0.68$, $p < 0.05$); and severely stenotic regions (>70%) demonstrated lower consistency than mild-to-moderate stenotic regions (75.4% vs. 83.6%). These factors collectively contribute to the observed higher consistency in proximal major vessels compared to distal small branches, as visually evident in Figure 18.

Through comprehensive morphological analysis and expert assessment, we verified the authenticity and accuracy of our generated dynamic virtual angiography images compared to clinical XA images. The overall morphological consistency of 80.9% and expert ratings averaging 4.2/5.0 for clinical diagnostic value demonstrate the potential clinical utility of this approach. However, it is important to note that X-ray angiography and 4D-CTA represent two distinct physical imaging systems requiring separate scanning procedures, during which patient positioning and physiological states may differ. These factors, along with vessel size, projection angles, cardiac phase, image noise, and stenosis severity, contribute to the observed discrepancies between actual XA images and simulated XA images, particularly in smaller distal branches. Such variations are expected and should be considered when interpreting virtual angiography results in clinical settings.

*3.2. Consistency Validation of Coronary Guidewire Range of Motion*

This section aims to validate the consistency between the range of motion of coronary guidewires in our dynamic cardiovascular phantom and in real clinical interventional procedures. Based on the clinical data acquisition and dynamic phantom experiments described previously, we compared the spatial dynamic characteristics of guidewires in actual coronary interventions with those in phantom experiments.

Figure 22 presents a comparison between real clinical XA images and dynamic phantom experimental results. The upper row displays a coronary guidewire inserted into the RCA during an actual clinical interventional procedure, with two representative frames from the angiography sequence shown in the dashed boxes. The rightmost image presents the superimposition of guidewire positions from all frames, visually demonstrating the complete spatial range of motion during the actual intervention. The lower row shows results using the same guidewire model in our dynamic phantom, with three-dimensional guidewire trajectories reconstructed via a dual-camera system and superimposed onto virtual angiographic images, displaying the range of motion within the phantom environment.

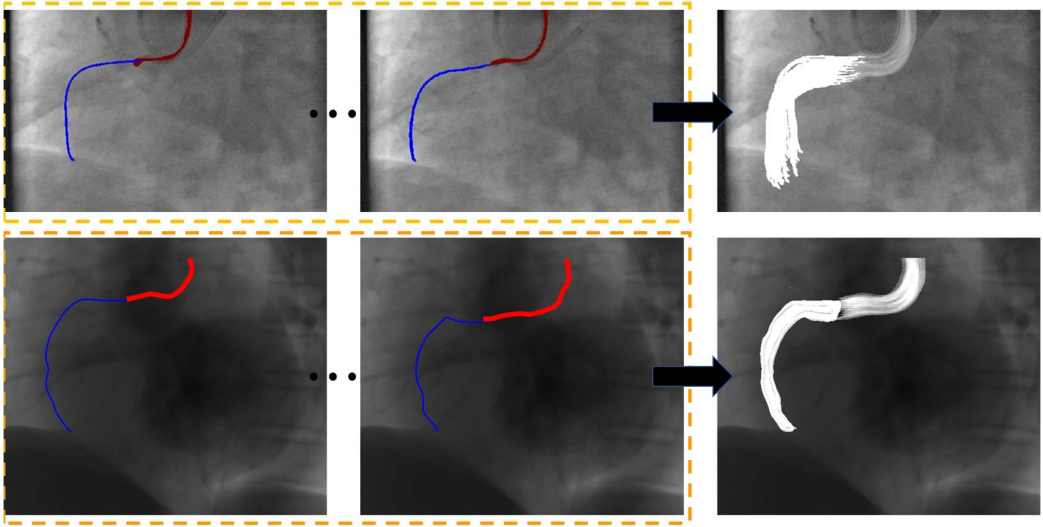

Figure 22: Clinical Coronary Guidewire Motion Versus Dynamic Phantom Experimental Comparison, with blue lines representing the Coronary Guidewire and red lines indicating the 6F catheter.

### 3.2.1. Standard Angiographic Projections for Guidewire Motion Analysis

To evaluate guidewire behavior comprehensively, we selected four standard angiographic projections commonly used in clinical practice: (1) LAO 34.3°/CRA 29.7°: Provides optimal visualization of coronary ostia and proximal segments. (2) RAO 30.2°/CRA 0.2°: Horizontal right anterior oblique view for mid-segments with minimal foreshortening. (3) RAO 32.4°/CAU 0.3°: Similar to previous angle but with slight caudal angulation. (4) RAO 32.4°/CAU 32.1°: Steep RAO-caudal view highlighting distal segments and branches.

### 3.2.2. Quantitative Analysis Results

We employed multiple mathematical metrics to quantitatively assess the consistency of guidewire range of motion across all four standard projection angles in both left and right coronary systems.

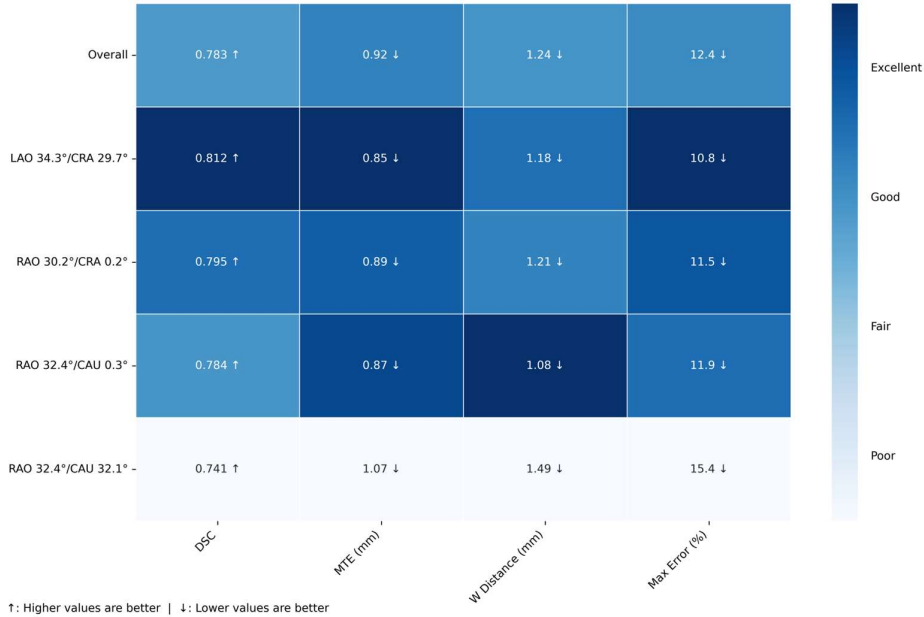

Figure 23: Visualization of Guidewire Motion Consistency Metrics.

The quantitative assessment of guidewire motion consistency employs four complementary mathematical metrics: Dice Similarity Coefficient ( $\text{DSC} = \frac{2|X \cap Y|}{|X|+|Y|}$ ) measures spatial overlap between binary trajectory masks $X$ and $Y$; Mean Trajectory Error ( $\text{MTE} = \frac{1}{N}\sum_{i=1}^{N}\|p_i - q_i\|_2$ ) quantifies average Euclidean distance between corresponding trajectory points $p_i$ and $q_i$; Wasserstein Distance ( $W(P,Q) = \inf_{\gamma \in \Gamma(P,Q)} \int_{M \times M} d(x,y) d\gamma(x,y)$ ) calculates optimal transport cost between trajectory distributions, where $\gamma$ denotes a coupling between distributions $P$ and $Q$, $\Gamma(P,Q)$ is the set of all possible couplings, $M \times M$ represents the Cartesian product of the metric space $M$ where the distributions are defined, $d(x,y)$ is the distance function, and $d\gamma(x,y)$ is the measure under coupling $\gamma$ (representing mass transported from $x$ to $y$); Maximum Error ( $\text{ME} = \max_i \|p_i - q_i\|_2 / L \times 100\%$ ) represents worst-case deviation as percentage, where $L$ is the total guidewire length.

The visualization in Figure 23 demonstrates that our dynamic phantom achieves high consistency with clinical data across multiple metrics and projection angles. The Dice Similarity Coefficient shows strongest agreement in the LAO 34.3°/CRA 29.7° projection (0.812), while the RAO 32.4°/CAU 32.1° view presents the greatest challenge (0.741). Mean Trajectory Error remains below 1.1 mm across all projections, with the steepest RAO-caudal view showing slightly higher error values.

Overall, the quantitative analysis results across all four standard angiographic projections demonstrate that our developed dynamic cardiovascular phantom effectively simulates the spatial activity characteristics of coronary guidewires in real interventional procedures for both left and right coronary systems. This provides a reliable experimental platform for interventional surgery simulation training, novel interventional device evaluation, and preoperative planning.

### 3.3. CABG End-to-Side Anastomosis Training on Dynamic Cardiac Phantom

To master the technical skills required for CABG procedures, extensive practice in a controlled environment is necessary before clinical application. End-to-side anastomosis represents a particularly challenging aspect of CABG surgery, requiring precise hand-eye coordination and meticulous suturing technique. To address this training need, we have made preliminary attempts to develop a simulation platform utilizing our dynamic cardiac phantom technology.

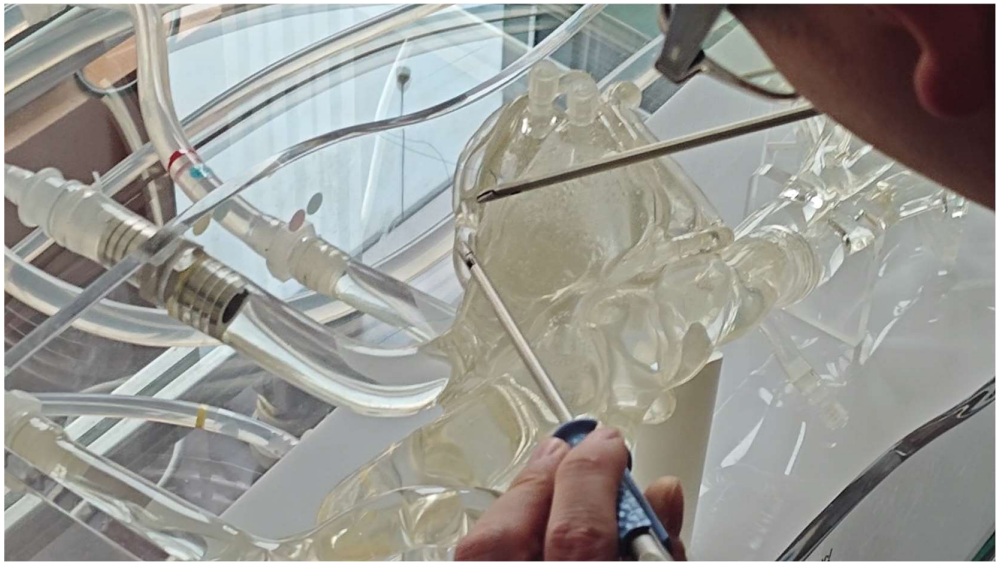

Figure 24: Close-up view of a transparent cardiac phantom during an experimental procedure.

Figure 24 shows a coronary artery bypass grafting surgical suturing training scenario on a transparent cardiac phantom. On this high-fidelity dynamic cardiac physical-digital twin model, the physician is performing an end-to-side anastomosis—one of the most critical technical procedures in coronary artery bypass surgery. The operator can be seen using precise surgical instruments (needle holder and forceps) to perform accurate suturing on the transparent model, simulating the process of connecting a graft vessel to a coronary artery. The complete operational procedure can be viewed in our supplementary materials.

This transparent cardiac phantom-based training system is manufactured using medical-grade silicone elastomers with hardness precisely controlled between 30-40A to simulate the elastic compliance of real coronary vessels. The model not only replicates the anatomical structures found in real surgical environments but is also equipped with functions simulating cardiac pulsation and blood flow, providing surgeons with tactile and visual feedback closely resembling actual surgery. Through this high-fidelity simulation training, surgeons can repeatedly practice refined suturing techniques in a risk-free environment, mastering proper needle spacing, suturing angles, and tension control, thereby accumulating valuable experience for clinical operations.

## 4. Discussion

This study presents a comprehensive framework for developing patient-specific dynamic cardiac models that integrate digital twin technology with physical simulation systems. Our approach addresses several critical gaps in current interventional cardiology training and planning methodologies. The following discussion examines the significance of our findings, compares them with existing literature, acknowledges limitations, and proposes future research directions.

*4.1. Significance of Patient-Specific Dynamic Cardiac Models*

The development of patient-specific dynamic cardiac models represents a significant advancement in personalized medicine. Our framework successfully transforms static 4D-CTA data into comprehensive dynamic models that accurately capture both anatomical structures and physiological motion patterns. This transformation is particularly valuable for complex coronary interventions where understanding patient-specific vascular dynamics is crucial for procedural success.

The high consistency (80.9%) between our virtual coronary angiography images and real XA images demonstrates the clinical validity of our approach. This level of fidelity is comparable to or exceeds that reported in previous studies [26,27], which demonstrated significant consistency between simulated and real angiographic images. The expert assessment results further validate the clinical utility of our models, particularly for main coronary vessels where ratings exceeded 4.3/5.0.

The integration of cardiac output analysis with dynamic coronary models provides a more comprehensive understanding of cardiovascular function than traditional static models. By quantifying parameters such as stroke volume, ejection fraction, and aortic valve dynamics, our system offers valuable insights into both anatomical and functional aspects of cardiac performance. This integrated approach aligns with recent trends in cardiovascular imaging that emphasize the importance of functional assessment alongside anatomical visualization [28,29].

*4.2. Advantages of the Digital-Physical Twin Approach*

Our digital-physical twin approach offers several advantages over purely virtual or purely physical simulation systems. By combining high-fidelity 3D digital models with transparent silicone physical phantoms, we provide both visual and tactile feedback that closely mimics real interventional procedures. This dual-modality approach addresses a significant limitation of existing training systems, which typically excel in either visual or tactile simulation but rarely both [17,18].

The quantitative validation of guidewire motion characteristics demonstrates that our physical phantoms accurately reproduce the spatial dynamics observed in clinical procedures. With Dice Similarity Coefficients ranging from 0.741 to 0.812 across different projection angles, our system demonstrates fidelity comparable to previously reported physical simulation systems [31,32]. This suggests that our approach effectively reproduces clinically relevant guidewire dynamic characteristics.

Furthermore, our approach to dynamic virtual angiography generation enables real-time visualization from multiple projection angles without additional radiation exposure. This capability is particularly valuable for training scenarios where repeated imaging would be impractical or unsafe in clinical settings. The ability to synchronize physical device manipulation with virtual fluoroscopic visualization creates an immersive training environment that closely resembles the clinical workflow.

*4.3. Educational and Clinical Applications*

The educational value of our system is demonstrated through its application in CABG end-to-side anastomosis training. The transparent nature of our physical models allows direct visualization of suturing techniques, while the dynamic capabilities simulate the challenges of operating on a beating heart. This combination provides a unique training platform that addresses the limitations of current static models used for surgical training [17].

From a clinical perspective, our framework has potential applications in preoperative planning, device selection, and procedural rehearsal. The ability to create patient-specific models that accurately reflect both anatomical and dynamic characteristics allows clinicians to anticipate challenges and optimize approaches before entering the catheterization laboratory. This capability is particularly valuable for complex cases involving severe stenosis, bifurcation lesions, or anomalous anatomy [33,34].

The binocular stereo vision system for guidewire reconstruction further enhances the clinical utility of our framework by providing real-time 3D positional information without fluoroscopic imaging. This approach achieves sub-millimeter accuracy in guidewire localization, comparable to commercial electromagnetic tracking systems [12] but without the need for specialized equipment or modifications to standard interventional devices.

*4.4. Limitations and Future Directions*

Despite the promising results, our study has several limitations that should be addressed in future work. First, the validation was performed on a single patient case with three-vessel coronary artery disease. While this case provided comprehensive data for model development, additional validation across a diverse patient population would strengthen the generalizability of our findings.

Second, the physical properties of our silicone models, while carefully controlled for shore hardness (30-40A), may not fully replicate the complex biomechanical behavior of living tissues. Future iterations could incorporate advanced biomaterials with anisotropic properties that more closely mimic the layered structure of arterial walls [35,36].

Third, our current system focuses primarily on coronary interventions and CABG procedures. Expanding the framework to include other cardiovascular interventions, such as transcatheter valve replacements, electrophysiology procedures, or peripheral vascular interventions, would broaden its clinical applicability.

Future research directions should include:
(1) Integration of fluid dynamics simulation to model blood flow patterns and pressure gradients within the coronary vasculature, enhancing the physiological realism of the system.
(2) Development of automated segmentation algorithms specifically optimized for severely stenotic coronary arteries, reducing the need for manual annotation and improving workflow efficiency.
(3) Incorporation of haptic feedback systems that dynamically adjust resistance based on the simulated vessel characteristics and lesion morphology.
(4) Longitudinal studies to assess the impact of training with our system on clinical

performance and patient outcomes.

Exploration of augmented reality interfaces that overlay virtual information directly onto the physical models, further enhancing the mixed reality training experience.

*4.5. Conclusion*

Our patient-specific dynamic digital-physical twin framework represents a significant advancement in cardiovascular simulation technology. By integrating 4D-CTA-derived digital models with physical phantoms and virtual angiography, we have created a comprehensive system that accurately reproduces both the anatomical and dynamic characteristics of coronary vasculature. The high consistency between our simulated models and clinical data validates the approach and demonstrates its potential for interventional training and preoperative planning.

This framework addresses a critical gap in current training methodologies by providing a dynamic, patient-specific environment that combines visual and tactile feedback. As interventional cardiology continues to advance toward more personalized approaches, simulation systems that accurately reflect individual patient characteristics will become increasingly valuable for both education and clinical decision-making.

## 5. Acknowledgment

*5.1. Statements of Ethical Approval*
This study was conducted in accordance with the declaration of Helsinki. The Ethics Committee of the General Hospital of the Chinese People's Liberation Army accepted this protocol (S2024-580). Informed consent was obtained from all individual participants included in the study.

*5.2. Funding*
This study was funded by National Natural Science Foundation of China (62031020 and 62476286), National Key R&D Plan(2022YFB4700800), Beijing Science and Technology Plan Project(Z231100005923039).

## References

[1] M.J. Budoff, D. Dowe, J.G. Jollis, et al., Diagnostic performance of 64-multidetector row coronary computed tomographic angiography for evaluation of coronary artery stenosis in individuals without known coronary artery disease, J. Am. Coll. Cardiol. 52(21) (2008) 1724-1732. https://doi.org/10.1016/j.jacc.2008.07.031.

[2] W.Y. Kim, P.G. Danias, M. Stuber, et al., Coronary magnetic resonance angiography for the detection of coronary stenoses, N. Engl. J. Med. 345(26) (2001) 1863-1869. https://doi.org/10.1056/NEJMoa010866.

[3] P.J. Scanlon, D.P. Faxon, A.M. Audet, et al., ACC/AHA guidelines for coronary angiography, J. Am. Coll. Cardiol. 33(6) (1999) 1756-1824. https://doi.org/10.1016/s0735-1097(99)00126-6.

[4] P. Eslami, J. Seo, A.A. Rahsepar, et al., Computational study of computed tomography contrast gradients in models of stenosed coronary arteries, J. Biomech. Eng. 137(9) (2015) 091002. https://doi.org/10.1115/1.4030891.

[5] D. Zeng, E. Boutsianis, M. Ammann, et al., A study on the compliance of a right coronary artery and its impact on wall shear stress, J. Biomech. Eng. 130(4) (2008) 041014. https://doi.org/10.1115/1.2937744.

[6] F. Tao, M. Zhang, A.Y.C. Nee, Digital twin driven smart manufacturing, Academic Press, 2019. https://doi.org/10.1016/C2018-0-02206-9.

[7] J. Corral-Acero, F. Margara, M. Marciniak, et al., The 'Digital Twin' to enable the vision of precision cardiology, Eur. Heart J. 41(48) (2020) 4556-4564. https://doi.org/10.1093/eurheartj/ehaa159.

[8] S.A. Niederer, J. Lumens, N.A. Trayanova, Computational models in cardiology, Nat. Rev. Cardiol. 16(2) (2019) 100-111. https://doi.org/10.1038/s41569-018-0104-y.

[9] R. Chabiniok, V.Y. Wang, M. Hadjicharalambous, et al., Multiphysics and multiscale modelling, data-model fusion and integration of organ physiology in the clinic: ventricular cardiac mechanics, Interface Focus. 6(2) (2016) 20150083. https://doi.org/10.1098/rsfs.2015.0083.

[10] M. Annabestani, A. Olyanasab, B. Mosadegh, Application of Mixed/Augmented Reality in Interventional Cardiology, J. Clin. Med. 13(15) (2024) 4368. https://doi.org/10.3390/jcm13154368.

[11] J.N.A. Silva, M. Southworth, C. Raptis, J. Silva, Emerging applications of virtual reality in cardiovascular medicine, JACC Basic Transl. Sci. 3(3) (2018) 420-430. https://doi.org/10.1016/j.jacbts.2017.11.009.

[12] M. Annabestani, S. Sriram, A. Caprio, et al., High-fidelity pose estimation for real-time extended reality (XR) visualization for cardiac catheterization, Sci. Rep. 14 (2024) 26962. https://doi.org/10.1038/s41598-024-76384-z.

[13] S. Jang, G.Y. Jung, S.J. Kang, et al., Current status of augmented reality in cerebrovascular surgery: a systematic review, Neurosurg. Rev. 44(3) (2021) 1233-1247. https://doi.org/10.1007/s10143-022-01733-3.

[14] P. Li, X. Zhang, X. Hu, F. Ding, C. Liang, Human-computer interaction on virtual reality-based training system for vascular interventional surgery, Comput. Methods Programs Biomed. 265 (2025) 108731. https://doi.org/10.1016/j.cmpb.2025.108731.

[15] J. Gosai, M. Purva, J. Gunn, Simulation in cardiology: state of the art, Eur. Heart J. 36(13) (2015) 777-783. https://doi.org/10.1093/eurheartj/ehu527.

[16] T.R. Coles, D. Meglan, N.W. John, The role of haptics in medical training simulators: a survey of the state of the art, IEEE Trans. Haptics. 4(1) (2011) 51-66. https://doi.org/10.1109/TOH.2010.19.

[17] J.I. Fann, A.D. Caffarelli, G. Georgette, et al., Improvement in coronary anastomosis with cardiac surgery simulation, J. Thorac. Cardiovasc. Surg. 136(6) (2008) 1486-1491. https://doi.org/10.1016/j.jtcvs.2008.08.016.

[18] P.S. Ramphal, D.N. Coore, M.P. Craven, et al., A high fidelity tissue-based cardiac surgical simulator, Eur. J. Cardiothorac. Surg. 27(5) (2005) 910-916. https://doi.org/10.1016/j.ejcts.2004.12.049.

[19] F. Kong, S.C. Shadden, Learning whole heart mesh generation from patient images for computational simulations, IEEE Trans. Med. Imaging. 42(2) (2023) 533 – 545. https://doi.org/10.1109/TMI.2022.3219284.

[20] M.J. Cardoso, W. Li, R. Brown, N. Ma, E. Kerfoot, Y. Wang, B. Murrey, A. Myronenko, C. Zhao, D. Yang, V. Nath, MONAI: An open-source framework for deep learning in healthcare [preprint], arXiv:2211.02701, 2022. https://doi.org/10.48550/arXiv.2211.02701.

[21] D.H. Pak, M. Liu, T. Kim, L. Liang, A. Caballero, J. Onofrey, S.S. Ahn, Y. Xu, R. McKay, W. Sun, R. Gleason, J.S. Duncan, Patient-specific heart geometry modeling for solid biomechanics using deep learning, IEEE Trans. Med. Imaging. 43(1) (2024) 203–215. https://doi.org/10.1109/TMI.2023.3294128.

[22] S. Wang, T. Ren, N. Cheng, L. Zhang, R. Wang, Innovative Integration of 4D Cardiovascular Reconstruction and Hologram: A New Visualization Tool for Coronary Artery Bypass Grafting Planning [preprint], arXiv:2504.19401, 2025. https://doi.org/10.48550/arXiv.2504.19401.

[23] A. Updegrove, N.M. Wilson, J. Merkow, H. Lan, A.L. Marsden, S.C. Shadden, SimVascular: An open source pipeline for cardiovascular simulation [software], Ann. Biomed. Eng. 45(3) (2017) 525–541. https://doi.org/10.1007/s10439-016-1762-8.

[24] S. Wang, T. Ren, N. Cheng, R. Wang, L. Zhang, Time-Varying Coronary Artery Deformation: A Dynamic Skinning Framework for Surgical Training [preprint], arXiv:2503.02218, 2025. https://doi.org/10.48550/arXiv.2503.02218.

[25] Z. Zhang, A flexible new technique for camera calibration, IEEE Trans. Pattern Anal. Mach. Intell. 22(11) (2002) 1330-1334. https://doi.org/10.1109/34.888718.

[26] C. Schumann, M. Neugebauer, R. Bade, et al., Implicit vessel surface reconstruction for visualization and simulation, Int. J. Comput. Assist. Radiol. Surg. 2(5) (2008) 275-286. https://doi.org/10.1007/s11548-007-0137-x.

[27] A.M. Vukicevic, S. Çimen, N. Jagic, et al., Three-dimensional reconstruction and NURBS-based structured meshing of coronary arteries from the conventional X-ray angiography projection images, Sci. Rep. 8 (2018) 1711. https://doi.org/10.1038/s41598-018-19440-9.

[28] G. Pontone, D. Andreini, A.I. Guaricci, et al., Incremental diagnostic value of stress computed tomography myocardial perfusion with whole-heart coverage CT scanner in intermediate- to high-risk symptomatic patients suspected of coronary artery disease, JACC Cardiovasc. Imaging. 12(2) (2019) 338-349. https://doi.org/10.1016/j.jcmg.2017.10.025.

[29] A.R. Ihdayhid, B.L. Norgaard, S. Gaur, et al., Prognostic value and risk continuum of noninvasive fractional flow reserve derived from coronary CT angiography, Radiology. 292(2) (2019) 343-351. https://doi.org/10.1148/radiol.2019182264.

[30] W.I. Willaert, R. Aggarwal, I. Van Herzeele, et al., Recent advancements in medical simulation: patient-specific virtual reality simulation, World J. Surg. 36(7) (2012) 1703-1712. https://doi.org/10.1007/s00268-012-1489-0.

[31] V. Luboz, Y. Zhang, S. Johnson, et al., ImaGiNe Seldinger: first simulator for Seldinger technique and angiography training, Comput. Methods Programs Biomed. 111(2) (2013) 419-434. https://doi.org/10.1016/j.cmpb.2013.05.014.

[32] C. Tercero, S. Ikeda, T. Uchiyama, et al., Autonomous catheter insertion system using magnetic motion capture sensor for endovascular surgery, Int. J. Med. Robot. 3(1) (2007) 52-58. https://doi.org/10.1002/rcs.116.

[33] J.M. Otton, R. Spina, R. Sulas, et al., Left atrial appendage closure guided by personalized 3D-printed cardiac reconstruction, JACC Cardiovasc. Interv. 8(7) (2015) 1004-1006. https://doi.org/10.1016/j.jcin.2015.03.015.

[34] B. Ripley, T. Kelil, M.K. Cheezum, et al., 3D printing based on cardiac CT assists anatomic visualization prior to transcatheter aortic valve replacement, J. Cardiovasc. Comput. Tomogr. 10(1) (2016) 28-36. https://doi.org/10.1016/j.jcct.2015.12.004.

[35] G. Biglino, P. Verschueren, R. Zegels, et al., Rapid prototyping compliant arterial phantoms for in-vitro studies and device testing, J. Cardiovasc. Magn. Reson. 15(1) (2013) 2. https://doi.org/10.1186/1532-429X-15-2.

[36] Z. Qian, K. Wang, S. Liu, et al., Quantitative prediction of paravalvular leak in transcatheter aortic valve replacement based on tissue-mimicking 3D printing, JACC Cardiovasc. Imaging. 10(7) (2017) 719-731. https://doi.org/10.1016/j.jcmg.2017.04.005.